\documentclass[letterpaper]{article} 
\usepackage{aaai2026}  
\usepackage{times}  
\usepackage{helvet}  
\usepackage{courier}  
\usepackage[hyphens]{url}  
\usepackage{graphicx} 
\usepackage{natbib}  
\usepackage{caption} 
\usepackage{algorithm}
\usepackage{algorithmic}

\usepackage{amsmath}
\usepackage{amsfonts}
\usepackage{amssymb}
\usepackage{tabularx}
\usepackage{booktabs}   
\usepackage{multirow}

\usepackage{subcaption}
\usepackage{newfloat}
\usepackage{listings}
\DeclareCaptionStyle{ruled}{labelfont=normalfont,labelsep=colon,strut=off} 
\floatstyle{ruled}
\newfloat{listing}{tb}{lst}{}
\floatname{listing}{Listing}
\nocopyright

\title{On the Superimposed Noise Accumulation Problem in \\ Sequential Knowledge Editing of Large Language Models}
\author{
    Ding Cao\thanks{Co-first authors.}, Yuchen Cai\footnotemark[1], Yuqing Huang, Xuesong He, Rongxi Guo, \\ Guiquan Liu\thanks{Corresponding author.}, Guangzhong Sun \\
}
\affiliations{
    University of Science and Technology of China\\

    \{caoding, caiyuchen, huangyuq, hxsustc, guorongxi\}@mail.ustc.edu.cn\\
    \{gqliu, gzsun\}@ustc.edu.cn
}

\usepackage{bibentry}

\begin{document}

\maketitle

\begin{abstract}
Sequential knowledge editing techniques aim to continuously update knowledge in large language models at low cost, preventing models from generating outdated or incorrect information. However, existing sequential editing methods suffer from a significant decline in editing success rates after long-term editing. Through theoretical analysis and experiments, our findings reveal that as the number of edits increases, the model's output increasingly deviates from the desired target, leading to a drop in editing success rates. We refer to this issue as the \textbf{superimposed noise accumulation problem}. Our further analysis demonstrates that the problem is related to the erroneous activation of irrelevant knowledge and conflicts between activated knowledge. Based on this analysis, a method named \textbf{DeltaEdit} is proposed that reduces conflicts between knowledge through dynamic orthogonal constraint strategies. Experiments show that DeltaEdit significantly reduces superimposed noise, achieving a 16.8\% improvement in editing performance over the strongest baseline.
\end{abstract}


\section{Introduction}

Large language Models encode vast amounts of knowledge acquired during pretraining \cite{brown2020language, petroni2019language, roberts2020much}. This makes them invaluable as knowledge bases across various applications \cite{thirunavukarasu2023large, alkhamissi2022review}. However, they are prone to generating inaccurate or outdated information \cite{de2021editing, mitchell2021fast}, necessitating continuous updates to maintain their accuracy and reliability. While fine-tuning offers a potential solution for knowledge updates, it is computationally expensive and risks catastrophic forgetting \cite{luo2024empiricalstudycatastrophicforgetting, mitchell2022memory, huang2025selfaug}. To overcome these limitations, knowledge editing techniques have emerged as an efficient alternative. They enable precise updates with minimal computational cost, while preserving the integrity of existing knowledge \cite{meng2023locatingeditingfactualassociations, yao2023editinglargelanguagemodels, wang2023knowledgeeditinglargelanguage, li2025mindbridge}. 

Current mainstream editing methods adopt the locate-then-edit paradigm \cite{meng2023locatingeditingfactualassociations, ma2024perturbationrestrainedsequentialmodelediting,gu2024modeleditingharmsgeneral}, which involves first identifying the most impactful model parameters \( W \) and then introducing update parameters \( \Delta \) to perform the desired edit. While these methods excel in single-edit tasks, real-world applications often demand multiple consecutive updates to accommodate rapidly evolving knowledge. This shift highlights the importance of sequential editing, which demands performing a series of edits while ensuring that all updated knowledge is accurately integrated and retained. Prior studies \cite{ gupta2024model} have shown that naively extending single-edit methods to sequential tasks can lead to reduced edit success rates and degradation in model performance. To address these issues, researchers have explored several critical factors \cite{ma2024perturbationrestrainedsequentialmodelediting, gu2024modeleditingharmsgeneral, fang2024alphaedit}. These efforts have led to the development of new methods specifically designed for sequential editing.

\begin{figure}[tb!]
    \centering
    \includegraphics[width=0.45\textwidth]{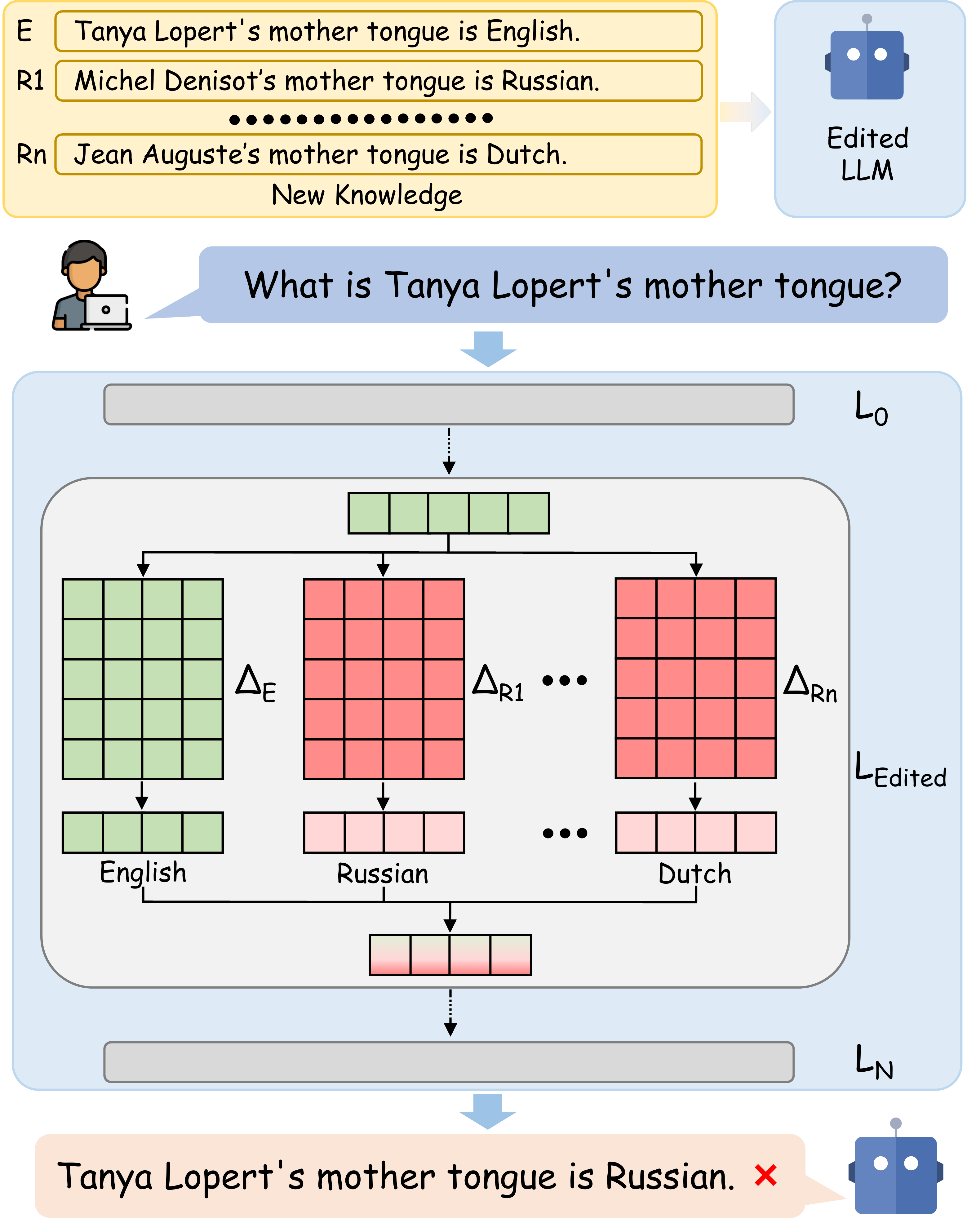} 
    \caption{Output deviation problem of edited LLM. The correctly activated output (green part) is interfered with by the incorrectly activated output (red parts).}
    \label{fig:first}
    \vskip -1.5em
\end{figure}

Despite recent advances, most existing studies on sequence editing remain superficial, they treat the update parameters for editing as a monolithic entity, neglecting the fine-grained dynamics of the update process and the potential interactions between successive edits. In this paper, we conduct a comprehensive investigation into the dynamic behavior of sequential editing. We identify a critical phenomenon: as the number of edits increases, the model's output increasingly deviates from its intended target. As shown in Figure \ref{fig:first}, a user query typically activates not only the correct knowledge but also numerous irrelevant knowledge. The superposition of irrelevant knowledge makes it difficult for the correct knowledge to be properly output. We refer to this issue as the superimposed noise accumulation problem. \textbf{Through our experiments, we demonstrate that the accumulation of superimposed noise is a major contributor leading to decreased editing success rates and even model collapse.}

To better understand the factors contributing to superimposed noise, we decompose the update parameter \(\Delta\) into the outer product of two components: influence vectors and activation vectors. Influence vectors determine the capacity of an update to modify the model’s output, whereas activation vectors control the extent to which updates are triggered by different inputs. Our analysis reveals that superimposed noise is primarily influenced by the incorrect activation of activation vectors caused by input representations, and the overlap of influence vectors during editing. \textbf{Existing methods primarily focus on optimizing activation vectors and often neglect influence vectors, leading to suboptimal updates.} This imbalance motivates the need for a more robust approach to sequential editing.

Based on the preceding analysis, we propose DeltaEdit, a novel sequential editing method designed to mitigate the effects of superimposed noise. DeltaEdit introduces a dynamic orthogonal constraint strategy to explicitly optimize influence vectors during the editing process, reducing interference between updates. \textbf{Experimental results demonstrate that, compared to existing methods, DeltaEdit more effectively mitigates superimposed noise and achieves superior sequential editing performance.} In particular, it attains a 16.8\% improvement in editing performance over the strongest baseline. Furthermore, DeltaEdit better preserves model capabilities and representation distribution, thereby significantly enhancing the reliability and sustainability of sequential editing. To summarize, the main contributions of this work are as follows: 

\begin{enumerate}
    \item We uncover and define superimposed noise as a core limitation in sequential editing tasks. Furthermore, through experiments on multiple models and editing methods, we demonstrate that the problem of superimposed noise accumulation is a critical cause of the decline in editing performance and even model collapse.

    \item We analyze the factors that contribute to superimposed noise. This understanding enables us to identify key mechanisms for reducing superimposed noise and enhancing the performance of sequential editing methods.

    \item We develop DeltaEdit, an innovative sequential editing method that incorporates a dynamic orthogonal constraint strategy. Through extensive experiments, we demonstrate that DeltaEdit significantly outperforms existing methods. 
\end{enumerate}

\section{Related Work}

\subsection{Knowledge Editing}
From the perspective of whether to modify model parameters, \citet{yao2023editinglargelanguagemodels} categorizes knowledge editing methods into two major types: parameter-preserving methods and parameter-modifying methods. This paper primarily focuses on the latter. One line of work employs meta-learning approaches to edit language models through a hypernetwork. KE \cite{decao2021editingfactualknowledgelanguage} utilizes a bidirectional LSTM to predict weight updates for editing, whereas MEND \cite{mitchell2021fast} applies a low-rank decomposition of gradients to fine-tune the language model. Another line of work, based on conclusions drawn from causal probes \cite{dai2022knowledgeneuronspretrainedtransformers, meng2023locatingeditingfactualassociations}, performs edits in the feed-forward networks of middle model layers. KN \cite{dai2022knowledgeneuronspretrainedtransformers} achieves knowledge editing by modifying the activation values of specific neurons. ROME \cite{meng2023locatingeditingfactualassociations} uses normal equations to compute the update parameters required for editing, and MEMIT \cite{meng2023masseditingmemorytransformer} further extends this approach to support batch editing. AnyEdit \cite{jiang2025anyedit} effectively enables the editing of long-form knowledge.

\subsection{Sequential Editing}
\citet{gupta2024model} points out that performing multiple consecutive edits can lead to model performance degradation. \citet{yang2024butterfly} identified perplexity as an effective metric for detecting model collapse during sequence editing. By analyzing update parameters,  \citet{hu2024wilkewiselayerknowledgeeditor} finds that the overlap of input representations in the whitening space is a key factor contributing to poor editing performance. \citet{ma2024perturbationrestrainedsequentialmodelediting} theoretically analyzes that the bottleneck limiting sequential editing in models lies in the condition number of matrices and proposes PRUNE, which supports sequential editing by controlling the growth of the condition number. \citet{gu2024modeleditingharmsgeneral} observes that editing causes excessive parameter changes and proposes RECT, which improves editing performance by sparsifying update parameters. \citet{cai2024edit} proposed O-Edit, which reduces interference between successive updates by orthogonalizing the direction of each knowledge update. \citet{fang2024alphaedit} proposes AlphaEdit, which achieves nearly lossless sequential editing by restricting the solution space of update parameters to a specific null space.

\section{Analysis on Sequential Editing}
\subsection{Preliminary}

\textbf{Autoregressive Language Model.} Autoregressive language models \cite{minaee2024large} generate the next token based on the preceding tokens. Modern autoregressive models are predominantly composed of multiple stacked transformer layers. In such models, the hidden state of the $l$ layer, denoted as $h^l$, is calculated as:

{
\footnotesize
\begin{align}
    h^l &= h^{l-1} + a^l + m^l \\
    m^l &= W_{out}^l \sigma \left( W_{in}^l \gamma(h^{l-1} + a^l) \right)
\end{align}
}

Here, \( a_l \) is the output of the attention block, \( m_l \) is the output of the feed-forward network (FFN), \( \sigma \) is the activation function, and \( \gamma \) denotes layer normalization. Studies \cite{geva2021transformerfeedforwardlayerskeyvalue} show that the parameter matrices of the FFN, \( W_{\text{out}} \) and \( W_{\text{in}} \), encode knowledge acquired during pretraining. Building on this finding, most knowledge editing methods focus on modifying the FFN to update knowledge.

\textbf{Sequential Editing.}  Sequential editing \cite{wang2024wiserethinkingknowledgememory} refers to performing continuous model editing on a model. Model editing aims to modify the knowledge encoded in a model. Following previous research \cite{meng2023locatingeditingfactualassociations}, we define knowledge in the form of triples, represented as \( (s, r, o) \), where \( s \) is the subject, \( r \) is the relation, and \( o \) is the object. For example, if a model memorizes the knowledge triple \( (s = \text{iPhone}, r = \text{Latest model}, o = \text{iPhone 16}) \), providing a language prompt \( p(s, r) \), such as ``The latest model of iPhone is,'' as input causes the model to output ``iPhone 16.'' Model editing refers to the process of replacing the original knowledge triple \( (s, r, o) \) with a new knowledge triple \( (s, r, o^*) \), and we denote this operation as  \( \mathcal{E} = (s, r, o, o^*) \). In a sequential editing task of length \( T \), where an editing sequence \( \mathbb{E}_T = (\mathcal{E}_1, \mathcal{E}_2, \dots, \mathcal{E}_T) \) is performed, each edit \(\mathcal{E}_i\) builds upon the results of the previous edit \(\mathcal{E}_{i-1}\).

Current knowledge editing methods \cite{meng2023locatingeditingfactualassociations, meng2023masseditingmemorytransformer, fang2024alphaedit} primarily adopt the locate-then-edit paradigm to implement model editing. For a model \( M \), these methods first locate the most appropriate parameters \( W \) within the model, and then update them using the following approach:

{
\footnotesize
\begin{align}
    \Delta &= F(M, W, \mathcal{E}) \\
    W' &= W + \Delta 
\end{align}
}

\(F\) represents the function used to compute update parameter \(\Delta\) and \(W'\) denotes the updated parameters. Given editing sequence $\mathbb{E}_T$, the update involves a parameter sequence \( \mathcal{D}_T = (\Delta_1, \Delta_2, \dots, \Delta_T) \), where \( \Delta_{i+1} = F(M, W_i, \mathcal{E}_i) \) and \( W_{i+1} = W_i + \Delta_i \).

\subsection{Update Parameter}

Most knowledge editing methods solve for update parameter \(\Delta\) by applying the normal equation. For example, MEMIT \cite{meng2023masseditingmemorytransformer} leverages the knowledge storage properties of FFN layers, identifies the \( W_{out}^l \) of a suitable layer \(l\) as the target for editing, and proposes the following equation:

{
\footnotesize
\begin{align}
    \Delta  \triangleq \mathop{\arg\min}\limits_{\hat{\Delta}} \left( \| \hat{\Delta}  K_1 - R \|^2 + \| \hat{\Delta} K_0 \|^2 \right) 
\end{align}
}

According to this equation, the solution for $\Delta$ can be obtained in the following form:

{
\footnotesize
\begin{align}
    \Delta = R K_1^\top (C_0 + K_1 K_1^\top)^{-1}
\end{align}
}

where \( K_1 \) represents the input representation to \( W_{\text{out}}^l \) for the final token of \( s \) in the edited triplet \( (s, r, o^*) \). \( K_0 \) denotes the input representations for tokens that should remain unaffected. \( C_0 = \mathbf{E}(K_0 K_0^\top) \) is statistically estimated over a large dataset. \( R \), the learned editing representation, is obtained by optimizing the loss function:

\begin{equation}
loss = -\log P(o^* \mid M(p(s, r), R)) \label{eq:loss}
\end{equation}

AlphaEdit \cite{fang2024alphaedit} has a similar solution:
\begin{equation}
\Delta = R K_1^\top \mathbb{P} \left( K_p K_p^\top \mathbb{P} + K_1 K_1^\top \mathbb{P} + I \right)^{-1}
\end{equation}
where \( K_p \) denotes the input representations for tokens that have already been edited, and \( \mathbb{P} \) is the null space of \( K_0 \).

To simplify the discussion, we focus on editing one knowledge at a time. In this case, \(\Delta\) can be seen as the product of vector \(\alpha\) and the transpose of vector \(\beta\):

\begin{equation}
    \Delta = \alpha  \beta^\top, \alpha = R, 
\end{equation}
\begin{equation*}
    \beta =
    \begin{cases}
    (C_0 + K_1 K_1^\top)^{-1} K_1,  \text{MEMIT} \\
    \left(\mathbb{P} K_p K_p^\top + \mathbb{P} K_1 K_1^\top + I \right)^{-1} \mathbb{P} K_1,  \text{AlphaEdit}
    \end{cases}
\end{equation*}

We refer to \(\alpha\) as the influence vector and \(\beta\) as the activation vector. This naming is based on the following reasoning:

\begin{itemize}
    \item \(\alpha\) is a specially trained vector designed to modify the model's output.
    \item \(\beta\) determines the extent to which \(\alpha\) is activated. The calculation of \(\Delta k\) can be expressed as \((k^\top \beta) \alpha\), where the dot product of \(k\) and \(\beta\) determines the activation strength.
\end{itemize}

For all methods adopt a computation of \(\Delta\) similar to MEMIT or AlphaEdit, the resulting \(\Delta\) can similarly be decomposed as \( \alpha \beta^\top \) from a comparable perspective.

\begin{figure}[t]
    \centering
    \includegraphics[width=0.405\textwidth]{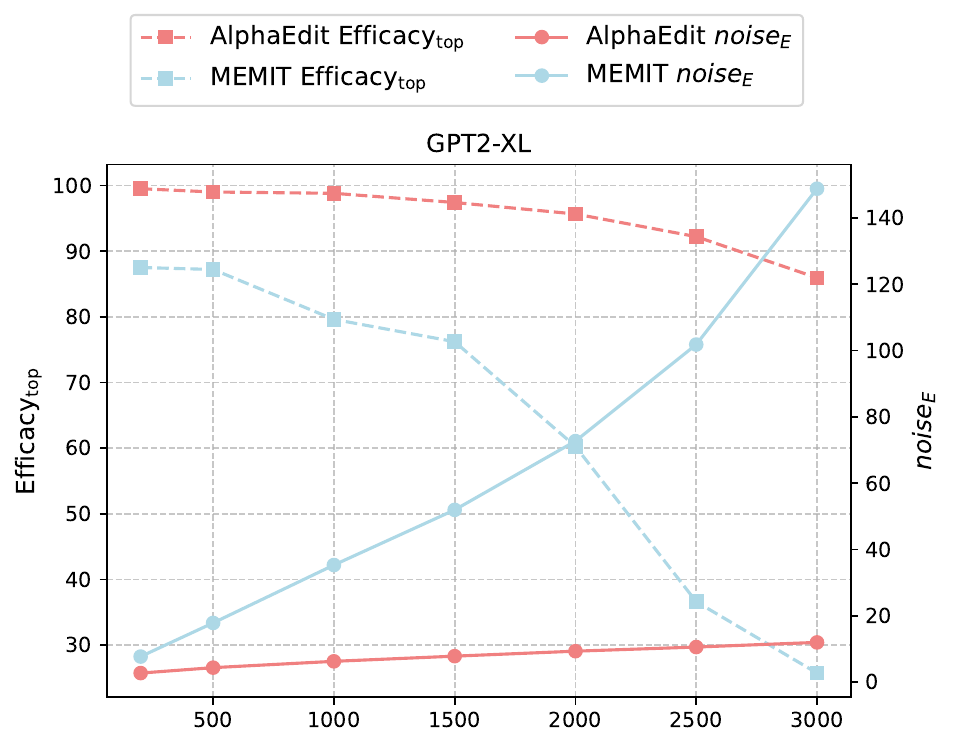}  
    \includegraphics[width=0.39\textwidth]{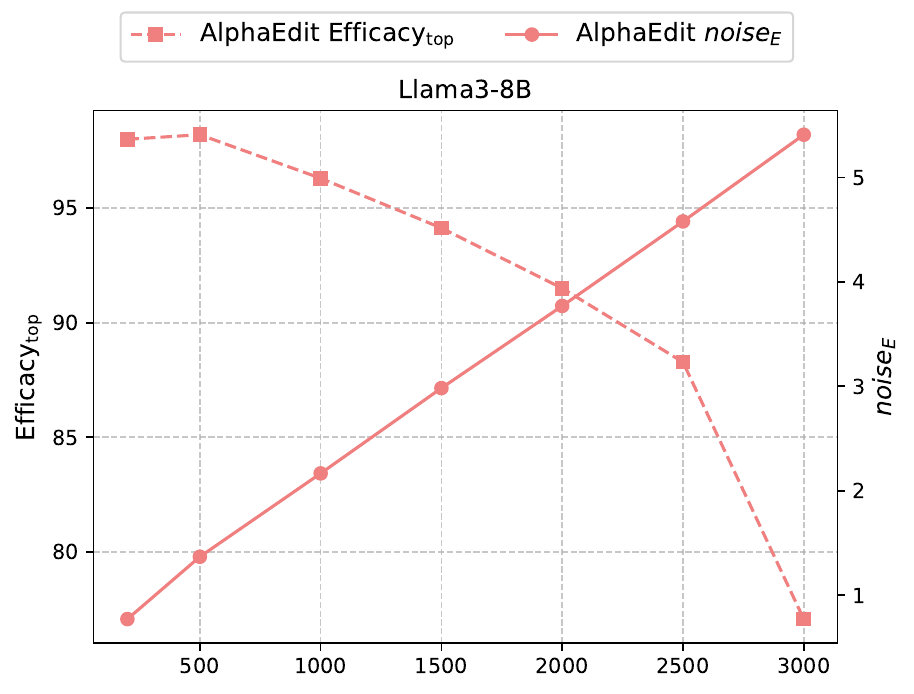}  
    \caption{The changes in Efficacy\textsubscript{top} and \( noise_E \) with the number of edits. The left figure displays results for GPT2-XL using AlphaEdit and MEMIT, the right figure shows results for LLaMA3-8B using only AlphaEdit. MEMIT is excluded from the figure of LLaMA3-8B due to its excessively high \( noise_E \), making visualization difficult.}
    \label{fig:noise_e}
\end{figure}

\subsection{The Problem of Superimposed Noise} \label{sec:analysis}

\textbf{Superimposed Noise.} In the sequential editing task, a series of editing operations is performed on the model. Upon completing the entire sequence of edits \( \mathbb{E}_T \), the parameters \( W_{out}^l \) of the model's \(l\)-th layer are updated. For the representation \( k_e \) of the subject token being edited during operation \( \mathcal{E}_e \), the output deviates from the original \( W_{out}^l k_e \). This deviation can be formally analyzed using the \( L_2 \) norm:

\begin{equation}
\| ( W_{out}^l + \sum_{i \leq T} \Delta_i ) k_e \|_2 \leq \| W_{out}^l k_e \|_2 + \| \sum_{i \leq T} \Delta_i k_e \|_2
\label{eq:deviation}
\end{equation}

Here, \( \| W_{out}^l k_e \|_2 \) is a constant, while \( \| \sum_i \Delta_i k_e \|_2 \) determines the upper bound of the deviation. In editing operation \( \mathcal{E}_e \), the update parameter can be expressed as \( \Delta_e \). Ideally, after multiple sequential edits, different editing operations should remain independent and not interfere with one another. Under such conditions, the deviation associated with \( \mathcal{E}_e \) satisfies:

\begin{equation}
\| \sum_{i \leq T} \Delta_i k_e \|_2^2 = \| \Delta_e k_e \|_2^2
\label{eq:ideal_noise}
\end{equation}

However, in practice, interference between different editing operations often occurs, leading to additional deviation in the output. To quantitatively characterize this phenomenon, we introduce the following definition.

\textbf{Definition 1.} The superimposed noise experienced by the editing operation \( \mathcal{E}_e \), denoted as \( noise_e \), is defined as
\begin{equation}
noise_e = \| \sum_{i \leq T} \Delta_i k_e \|_2^2 - \| \Delta_e k_e \|_2^2
\label{eq:practical_noise}
\end{equation}

This definition quantitatively describes the additional deviation, termed superimposed noise, which is induced by the interference among multiple editing operations. It provides a rigorous metric for evaluating the extent of such interference in sequential editing scenarios. To measure the overall superimposed noise level of all editing operations, we further define the average \( noise_E \) of all editing operations as:

\begin{equation}
noise_E = \frac{1}{T} \sum_{e \leq T} noise_e
\label{eq:noise}
\end{equation}

\textbf{The Impact of Noise on Editing Performance} \quad To investigate the impact of superimposed noise on editing performance, experiments are conducted using AlphaEdit and MEMIT on GPT2-XL \cite{Radford2019LanguageMA} and Llama3-8B \cite{dubey2024llama}. In these experiments, the models are sequentially edited using the CounterFact dataset \cite{meng2023locatingeditingfactualassociations}. Details of the experimental setup are provided in the Appendix \ref{sec:setup}. The results presented in Figure \ref{fig:noise_e} reveal the following trends:

\begin{itemize}
    \item \textbf{Performance Decline:} The model's editing performance decreases as the overall superimposed noise (\(noise_E\)) increases, indicating the detrimental impact of noise on performance.
    \item \textbf{Nonlinear Behavior:} The decline in performance is nonlinear. Once \(noise_E\) surpasses a certain threshold, the model's editing performance deteriorates sharply.
\end{itemize}

Moreover, as demonstrated in the Appendix \ref{sec:casestudy}, models edited by MEMIT with high levels of superimposed noise lose the ability to generate coherent outputs. \textbf{Unchecked accumulation of noise not only diminishes the efficacy of editing operations but can also substantially degrade overall model performance.} Therefore, mitigating superimposed noise is critical for ensuring robust and reliable model behavior.

\subsection{Factors Affecting Superimposed Noise} \label{sec:factor}
According to \( \Delta_e = \alpha_e \beta_e^\top \), the calculation of \( noise_e \) can be expanded into the following form:

\begin{equation}
noise_e = \sum_{\substack{i, j \\ (i, j) \neq (e, e)}} k_e^\top \beta_i \alpha_i^\top \alpha_j \beta_j^\top k_e
\label{eq:practical_noise_exp}
\end{equation}

By analyzing the computation of \( noise_e \), we find that its value is determined by two key factors: \( k_e^\top \beta_i \) and \( \alpha_i^\top \alpha_j \). The former is related to the incorrect activation of activation vectors, while the latter is related to the overlap of influence vectors during editing. Reducing either of these two terms can decrease \( noise_e \). 

Notably, existing methods such as MEMIT and AlphaEdit exhibit significant differences in editing performance. Given that AlphaEdit and MEMIT differ in their computation of activation vector \( \beta \), we hypothesize that the performance difference is due to AlphaEdit having fewer incorrect activations. To validate this hypothesis, we conduct experiments under the same settings as described in Section~\ref{sec:analysis}. Specifically, given a sequence of edits \( \mathbb{E}_T \), we compute the average \( k_e^\top \beta_i \) (\( i \neq e \)) across different numbers of edits, defined as:

\begin{equation}
k^\top \beta = \frac{1}{TN} \sum_{i \leq T} \sum_{j \neq i} k_i^\top \beta_j
\end{equation}

\begin{figure}[t]
    \centering
    \includegraphics[width=0.39\textwidth]{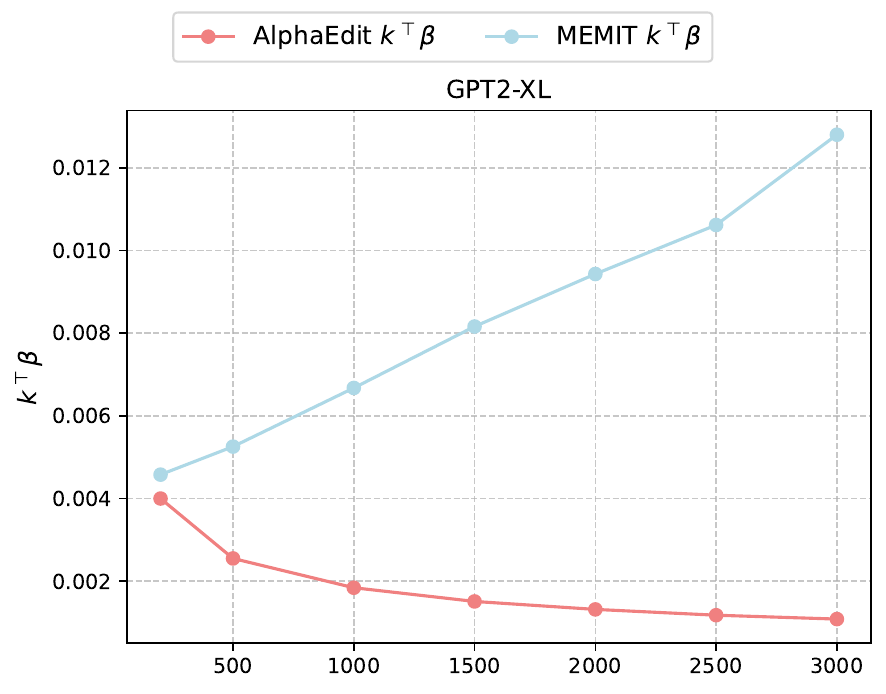}  
    \caption{The variation curve of \( k^\top \beta \) for AlphaEdit and MEMIT as the number of edits increases, with GPT2-XL serving as the edited model. As the results of Llama3-8B are not suitable for display in a line chart, they are presented in Appendix \ref{sec:sup-llama}}
    \label{fig:dots}
\end{figure}

The experimental results are shown in Figure \ref{fig:dots}. The results indicate that the \( k^\top \beta \) values obtained using AlphaEdit are significantly smaller than those of MEMIT. The improvement is primarily attributed to AlphaEdit's use of null space projection and the inclusion of \( K_p \) in the computation of \( \beta \), which reduces the overlap of information between \( \beta \) and the input representations of unrelated tokens. This demonstrates that optimizing \( \beta \) to minimize erroneous activations is an effective strategy. However, as the number of sequential edits increases, the performance of AlphaEdit still degrades significantly. The observation suggests that simply reducing \( k_e^\top \beta_i \) (\( i \neq e \)) is not sufficient to completely resolve the noise issue in sequential editing. 

\section{DeltaEdit}

The analysis in Section~\ref{sec:factor} reveals that to further reduce superimposed noise, greater attention should be given to the other influencing factor $\alpha_i^\top \alpha_j$. To ensure that $\alpha_i^\top \alpha_j$ (for all $i, j, i \neq j$) approaches zero, it is essential to minimize the overlap of \( \alpha \) with previous edits during training. Given the constrained solution space for \( \alpha \), we propose \textbf{DeltaEdit}, a method that leverages a dynamic orthogonal constraints strategy based on historical editing information to reduce interference while ensuring editing efficiency.

For a given sequence of edits $(\mathcal{E}_1, \mathcal{E}_2, \dots, \mathcal{E}_e)$, the noise $noise_e$ during the execution of the $e$-th edit operation $\mathcal{E}_e$ can be expressed as:

{
\footnotesize
\begin{align}
    noise_e &= \| \sum_{i \leq e} \Delta_i k_e \|_2^2  - \| \Delta_e k_e \|_2^2 \\
    &= \| \sum_{i < e} \Delta_i k_e \|_2^2 + 2 \sum_{i < e} k_e^\top \beta_e \alpha_e^\top \alpha_i \beta_i^\top k_e \nonumber
\end{align}
}

This formula shows that \( noise_e \) is determined by the cumulative interference from historical edits 
\(
(\| \sum_{i < e} \Delta_i k_e \|_2^2)
\) 
and the interaction between the current and historical edits 
\(
(\sum_{i < e} k_e^\top \beta_e \alpha_e^\top \alpha_i \beta_i^\top k_e)
\). 
It can be observed that the latter term can be reduced by constraining the training of \(\alpha\). 

\textbf{Orthogonal Constraint Strategy}

To suppress the growth of \( noise_e \), we leverage the accumulated parameters \( \Delta_{\text{history}} \) of historical edits when performing the current edit $\mathcal{E}_e$:

\begin{equation}
\Delta_{\text{history}} = \sum_{i < e} \Delta_i,
\end{equation}

And we introduce a dynamic threshold $t$. If $\| \Delta_{\text{history}} k_e \|_2^2 > t$, an orthogonal space projection optimization constraint is applied during the training of the influence vector $\alpha_e$ to control further growth of interference.

\textbf{Orthogonal Space Optimization}

To achieve orthogonality between the $\alpha_e$ and all vectors in the set $A = \{\alpha_i, i < e\}$ without storing these vectors, we compute a null space \cite{wang2021training} using singular value decomposition (SVD). Restricting the training of $\alpha_e$ to this null space ensures that $\alpha_e$ remains almost orthogonal to all vectors in $A$, effectively avoiding the storage overhead. To compute this null space, the column space matrix $D$ is first constructed:

\begin{equation}
D = \Delta_{\text{history}} \Delta_{\text{history}}^\top
\end{equation}

Since the column space dimension of $\Delta_{\text{history}}$ is smaller than its row space dimension, decomposing $D$ can produce the same column space as $\Delta_{\text{history}}$ while reducing the computational cost of the decomposition. Singular value decomposition (SVD) is then performed on $D$:

\begin{equation}
\{U, \Lambda, U^\top\} = \text{SVD}(D)
\end{equation}

Here, $\Lambda$ contains the eigenvalues, and $U$ contains the corresponding eigenvectors. The non-zero eigenvalues are collected into a set, denoted as $N$. To avoid excessively shrinking the training space, if the number of non-zero eigenvalues exceeds three-quarters of the dimension of $\alpha$, the smallest eigenvalues in $N$ are removed until the number of elements in $N$ equals three-quarters of $\alpha$'s dimension. The eigenvectors corresponding to the remaining eigenvalues in $N$ are selected to form $\hat{U}$. Using $\hat{U}$, the projection matrix $P$ that represents the null space is computed as:

\begin{equation}
P = I - \hat{U}\hat{U}^\top,
\end{equation}

where $I$ is the identity matrix. During the training of the $\alpha_e$, the optimizer updates $\alpha_e$, and after each update, $\alpha_e$ is projected onto the null space P:

\begin{equation}
\alpha_e = P \alpha_e.
\end{equation}

\begin{table*}[t!]
\centering
\begin{tabularx}{\textwidth}{llXXXXXXXXXXX}
\toprule
\multirow{2}{*}{\textbf{Model}} & \multirow{2}{*}{\textbf{Method}} & \multicolumn{3}{c}{\textbf{CounterFact\textsubscript{top}}} & \multicolumn{3}{c}{\textbf{CounterFact\textsubscript{larger}}} & \multicolumn{3}{c}{\textbf{ZsRE}} \\ 
\cmidrule(lr){3-5} \cmidrule(lr){6-8} \cmidrule(lr){9-11}
& & Eff.$\uparrow$ & Gen.$\uparrow$ & Spe.$\uparrow$ & Eff.$\uparrow$ & Gen.$\uparrow$ & Spe.$\uparrow$ & Eff.$\uparrow$ & Gen.$\uparrow$ & Spe.$\uparrow$ \\ 
\midrule
\multirow{6}{*}{\textbf{GPT2-XL}} 
    & FT & 27.17 & 8.3 & 2.22 & 69.93 & 50.43 & 18.24 & 5.13 & 4.43 & 0.28 \\
    & ROME & 0.53 & 0.38 & 0.41 & 53.93 & 49.55 & 84.56 & 44.81 & 41.12 & 7.52 \\
    & MEMIT & 25.73 & 16.85 & 12.71 & 67.80 & 60.85 & 68.99 & 26.8 & 25.38 & 13.51 \\
    & PRUNE & 7.93 & 6.45 & 8.91 & 60.00 & 56.13 & 80.94 & 2.94 & 2.77 & 3.49 \\
    & RECT & 56.77 & 31.13 & 18.11 & 88.3 & 74.8 & 72.85 & 39.5 & 36.13 & 15.5 \\
    & AlphaEdit & 85.93 & 46.23 & 53.18 & 98.37 & 88.05 & 92.73 & 93.19 & 84.97 & 23.55 \\
    & DeltaEdit & \textbf{93.8} & \textbf{54.37} & \textbf{53.28} & \textbf{98.97} & \textbf{91.1} & \textbf{92.91} & \textbf{95.26} & \textbf{88.48} & \textbf{24.96} \\
\midrule
\multirow{5}{*}{\textbf{Llama3-8B}} 
    & FT & 9.6 & 3.63 & 0.23 & 81 & 66.93 & 12.2 & 6.45 & 6.01 & 24.31 \\
    & ROME & 0 & 0 & 0 & 63.37 & 59.5 & 24.77 & 2.95 & 2.82 & 1.22 \\
    & MEMIT & 0 & 0 & 0 & 50.33 & 50.37 & \textbf{81.62} & 0 & 0 & 0 \\
    & PRUNE & 0 & 0 & 0 & 51.2 & 51.2 & 17.14 & 0 & 0 & 0 \\
    & RECT & 0 & 0 & 0 & 49.23 & 49.23 & 16.07 & 0 & 0 & 0 \\
    & AlphaEdit & 77.07 & 53.92 & 17.59 & 93.83 & 84.75 & 64.44 & 94.49 & \textbf{91.52} & 29.77 \\
    & DeltaEdit & \textbf{93.87} & \textbf{56.52} & \textbf{34.2} & \textbf{98.63} & \textbf{83.2} & 78.67 & \textbf{94.94} & 91.19 & \textbf{30.08} \\

\bottomrule
\end{tabularx}
\caption{Eff., Gen., and Spe. under \textbf{CounterFact\textsubscript{top}} and \textbf{ZsRE} denote \textbf{Efficacy\textsubscript{top}}, \textbf{Generalization\textsubscript{top}}, and \textbf{Specificity\textsubscript{top}}, while Eff., Gen., and Spe. under \textbf{Counterfact\textsubscript{larger}} denote \textbf{Efficacy\textsubscript{larger}}, \textbf{Generalization\textsubscript{larger}}, and \textbf{Specificity\textsubscript{larger}}. ↑ indicates that higher values are better. The bolded numbers are the largest for the corresponding metrics.}
\label{tab:performance}
\end{table*}

\textbf{Dynamic Threshold Design}

Since $\| \Delta_{\text{history}} k_e \|_2^2$ increases continuously as the number of edits grows, a fixed threshold is unsuitable. To address this, we introduce a sliding average strategy to dynamically update the threshold. Specifically, the mean $m$ and variance $v$ of $\| \Delta_{\text{history}} k_e \|_2^2$ are updated using a sliding average after each edit. The sliding average updates are defined as:

{
\footnotesize
\begin{align}
m^{(t+1)} &= \delta m^{(t)} + (1 - \delta) \| \Delta_{\text{history}} k_e \|_2^2, \\
v^{(t+1)} &= \delta v^{(t)} + (1 - \delta) \left( \| \Delta_{\text{history}} k_e \|_2^2 - m^{(t+1)} \right)^2
\end{align}
}

where $\delta \in [0, 1]$ is a sliding average coefficient that controls the balance between historical stability and sensitivity to recent updates. Based on these updates, the dynamic threshold $t$ is defined as:

\begin{equation}
t = m + \eta \sqrt{v},
\end{equation}

The hyperparameter $\eta$ determines the strength of the constraint.

\section{Experiments}

In this section, we conduct a comprehensive evaluation of DeltaEdit. First, we assess its performance in sequential editing tasks. Subsequently, we evaluate its effectiveness in mitigating $noise_E$ and its ability to maintain the model's hidden representations.

\subsection{Experiment Setup}

In experiments, we adopt the computation method from AlphaEdit to calculate $\beta$. The experiments are conducted on two language models: \textbf{GPT2-XL} \cite{Radford2019LanguageMA} and \textbf{Llama3-8B} \cite{dubey2024llama}. To comprehensively evaluate the effectiveness of DeltaEdit, we compare its performance with several baseline methods, including \textbf{Fine-Tuning} \cite{zhu2020modifying}, \textbf{ROME} \cite{meng2023locatingeditingfactualassociations}, \textbf{MEMIT} \cite{meng2023masseditingmemorytransformer}, \textbf{PRUNE} \cite{ma2024perturbationrestrainedsequentialmodelediting}, \textbf{RECT} \cite{gu2024modeleditingharmsgeneral}, and \textbf{AlphaEdit} \cite{fang2024alphaedit}. 

The evaluation is conducted on two widely recognized benchmark datasets: \textbf{ZsRE} \cite{levy-etal-2017-zero} and \textbf{CounterFact} \cite{meng2023locatingeditingfactualassociations}. Following prior work \cite{meng2023masseditingmemorytransformer}, for ZsRE, we employ Efficacy\textsubscript{top}, Generalization\textsubscript{top}, and Specificity\textsubscript{top} as evaluation metrics. For CounterFact, we employ Efficacy\textsubscript{larger}, Generalization\textsubscript{larger}, and Specificity\textsubscript{larger}. Additionally, for more comprehensive evaluation, Efficacy\textsubscript{top}, Generalization\textsubscript{top}, and Specificity\textsubscript{top} are also employed. Metrics with the subscript ``top'' focus on the token with the highest output probability, while metrics with the subscript ``larger'' focus on the relative probability of the target token. Detailed definitions are provided in the Appendix \ref{sec:metrics}. The experiments assess the performance of various methods after 3,000 sequential edits, highlighting the models' abilities in editing efficiency, semantic generalization, and preserving unedited content.

\subsection{Results of Experiment}

Table \ref{tab:performance} presents the performance of different editing methods. A detailed analysis of the experimental results is provided below.

\textbf{DeltaEdit Demonstrates Superior Editing Performance.} DeltaEdit demonstrates superior performance across most evaluation metrics compared to all baseline methods, underscoring its effectiveness in handling complex sequential editing tasks. Its performance is particularly strong on the CounterFact dataset. On the Llama3-8B model, DeltaEdit achieves \textbf{16.8\%}, \textbf{2.6\%}, and \textbf{16.61\%} higher scores than AlphaEdit in Efficacy\textsubscript{top}, Generalization\textsubscript{top}, and Specificity\textsubscript{top}, respectively. On the GPT2-XL model, the improvement of DeltaEdit over AlphaEdit is smaller. This is because AlphaEdit already performs well on GPT2-XL, leaving less room for improvement. However, DeltaEdit still achieves better overall results. On the ZsRE dataset, DeltaEdit does not show a significant advantage. The lower similarity between object in ZsRE reduces the impact of superimposed noise during sequential editing. This suggests that DeltaEdit's noise control mechanisms are particularly effective in scenarios with high object similarity, as seen in the CounterFact dataset.

\begin{figure}[tb!]
    \centering
    \includegraphics[width=0.405\textwidth]{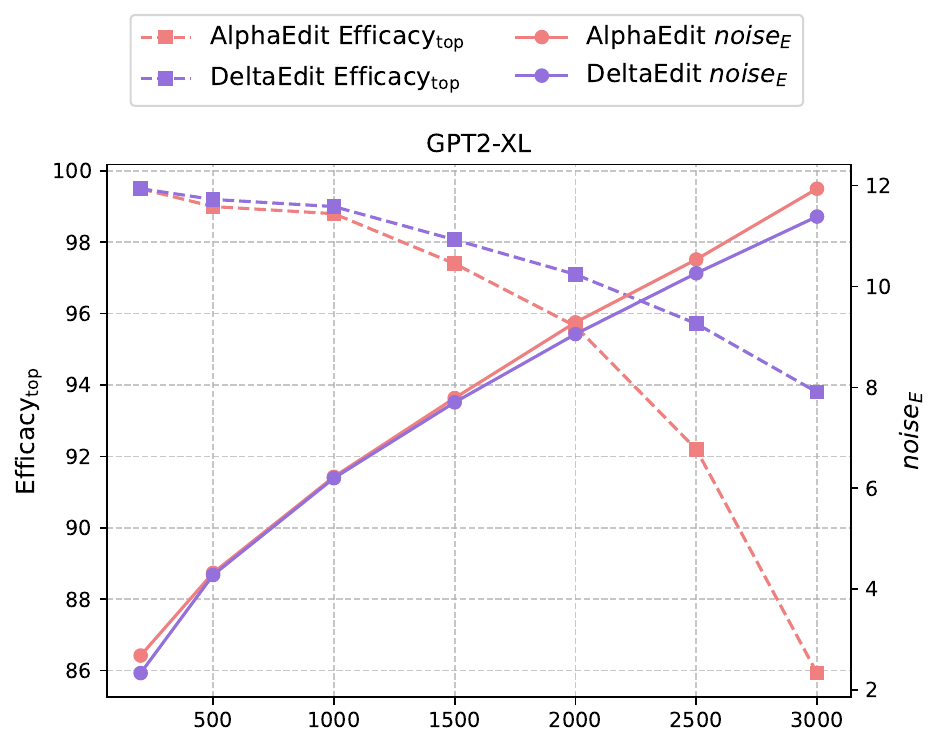}  
    \includegraphics[width=0.39\textwidth]{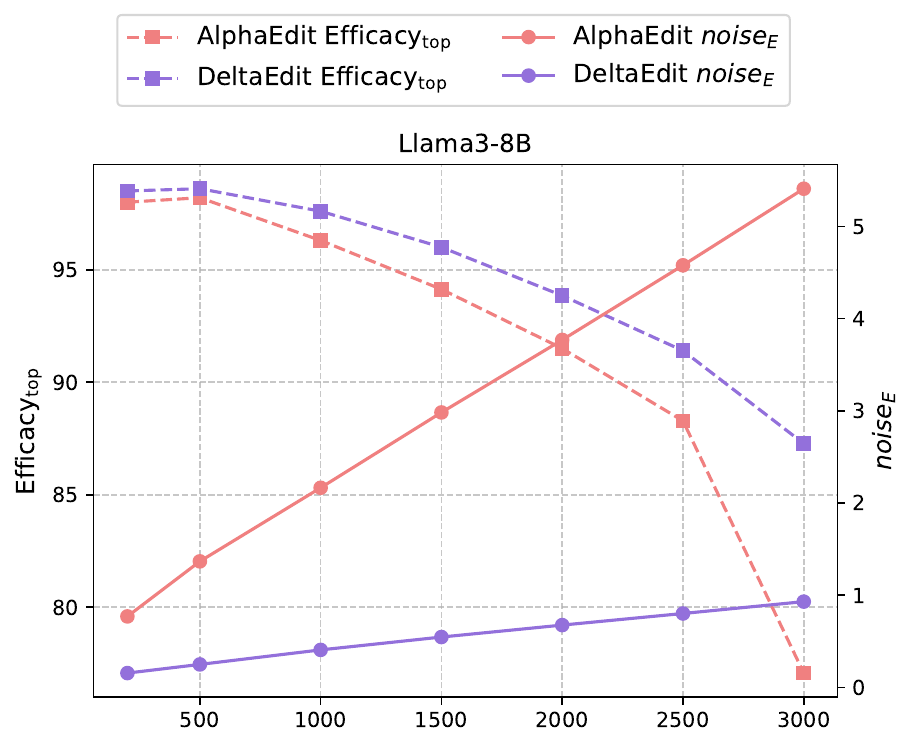}  
    \caption{The changes in Efficacy\textsubscript{top} and \( noise_E \) with the number of edits. The top figure shows results for GPT2-XL, the bottom figure shows results for LLaMA3-8B.}
    \label{fig:delta_noise_e}
    \vskip -1.5em
\end{figure}

\textbf{DeltaEdit Enhances Stability in Sequential Editing.} Sequentially performing 3,000 edits is an extremely challenging task, as frequent parameter updates can introduce more superimposed noise and degrade model performance over time. This issue is particularly evident in baseline methods, which perform poorly in this setting, especially on the Llama3-8B model. Despite its larger size, Llama3-8B has smaller parameter value ranges, making it more sensitive to noise accumulation. In contrast, DeltaEdit demonstrates remarkable stability and robustness during sequential editing. It maintains consistently strong performance, even after 3,000 consecutive edits, and significantly outperforms all baseline methods. These results highlight the effectiveness of DeltaEdit's noise control strategy, which effectively suppresses noise growth to maintain high accuracy and stability in long-term editing tasks.

\subsection{The Impact on Superimposed Noise}

Figure \ref{fig:delta_noise_e} illustrates the changes in Efficacy\textsubscript{top} and \( noise_E \) for GPT2-XL and Llama3-8B on the CounterFact dataset as the number of edits increases. \textbf{As shown in these figures, DeltaEdit effectively reduces \( \textbf{\textit{noise}}_\textbf{\textit{E}} \) while maintaining the stability of Efficacy\textsubscript{top}.} This observation further reinforces the empirical support for the hypothesis that superimposed noise has a significant impact on the effectiveness of editing methods. Furthermore, it confirms that reducing superimposed noise can substantially enhance editing performance.

A closer analysis reveals that in GPT2-XL, while the reduction in \( noise_E \) achieved by DeltaEdit is relatively modest, it significantly mitigates the decline in Efficacy\textsubscript{top}. This suggests that DeltaEdit can enhance the stability of editing performance even when the reduction in noise is limited. For Llama3-8B, the accumulation of noise within AlphaEdit leads to a significant decline in editing performance. In contrast, DeltaEdit exhibits a more notable impact, achieving a substantial reduction in \( noise_E \) while yielding a marked improvement in editing performance. This indicates that DeltaEdit exhibits superior optimization capabilities in noise-sensitive models.

In summary, the experimental results demonstrate that DeltaEdit substantially reduces superimposed noise, thereby enhancing the editing performance of models. These findings highlight the critical role of noise reduction in improving the stability and success of model editing.

\begin{figure}[tb!]
    \centering
    \includegraphics[width=0.43\textwidth]{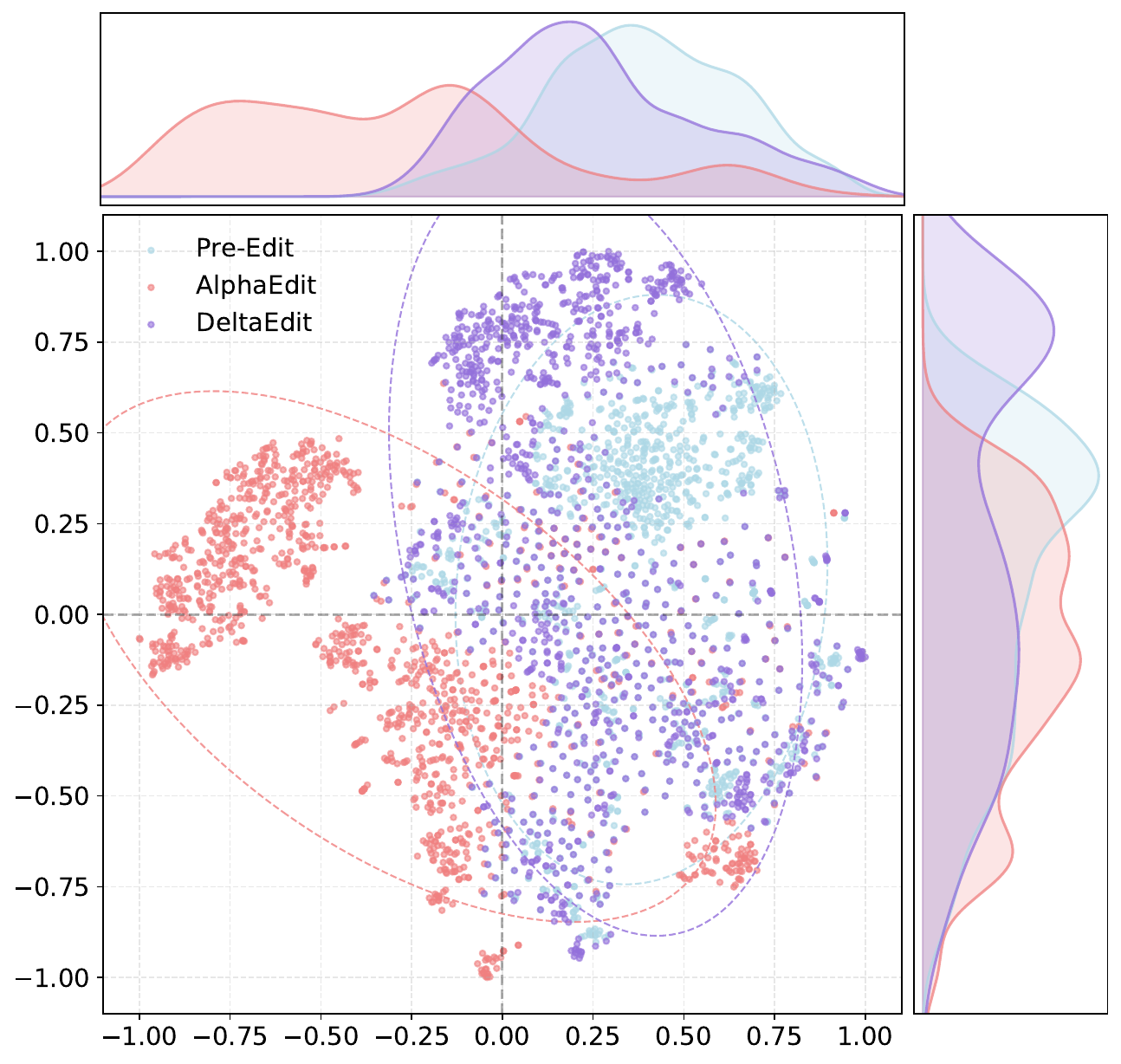}  
    \caption{The distribution of hidden representations of pre-edited and post-edited Llama3-8B after dimensionality reduction. The top and right curve graphs display the marginal distributions for two reduced dimensions. The dashed lines represent the 0.95 confidence intervals.}
    \label{fig:distribution}
    \vskip -1.5em
\end{figure}

\subsection{Hidden Representations Analysis} \label{sec:hidden}

Knowledge editing can alter a model's hidden representations. This study investigates these changes by comparing hidden representations from three Llama3-8B versions: pre-edited, edited with AlphaEdit, and edited with DeltaEdit (both edited 3,000 times on CounterFact). Representations were extracted using 1,000 factual prompts and visualized via t-SNE. \textbf{Results in Figure \ref{fig:distribution} indicate that DeltaEdit largely preserves the original distribution of hidden representations, unlike AlphaEdit, which causes significant shifts.} This minimal representational shift by DeltaEdit is crucial for maintaining the model's general capabilities.

\section{Conclusion}
In this work, we identify superimposed noise accumulation as a critical limitation in sequential editing of large language models. Comprehensive analysis demonstrates that this phenomenon leads to degraded editing performance and model collapse. Additionally, we trace this to the interference of correct knowledge by erroneously activated irrelevant knowledge. 

Based on our analysis, we propose DeltaEdit, which incorporates dynamic orthogonal constraints to mitigate superimposed noise by optimizing influence vectors that existing methods neglect. Extensive experiments show that DeltaEdit achieves superior sequential editing performance with 16.8\% improvement over strong baselines while better preserving model capabilities and representation integrity, thus providing a robust solution for reliable sequential knowledge editing.

\bibliography{aaai2026}

\newpage

\onecolumn

\appendix

\section{Experiment Setup}\label{sec:setup}

In this section, we provide detailed descriptions of experimental setup, including introduction to datasets, detailed explanation of evaluation metrics, implementation details, and discussion of baseline methods.

\subsection{Datasets}

\begin{itemize}
    \item \textbf{ZsRE} \cite{levy-etal-2017-zero} is a question answering dataset that uses questions generated through back-translation as equivalent neighboring questions. According to previous studies, Natural Questions are used as out-of-scope data to evaluate locality. Each sample in ZsRE includes a subject string and an answer as the editing target to evaluate editing success, along with paraphrased questions for generalization evaluation and locality questions for specificity evaluation.
    \item \textbf{Counterfact} \cite{meng2023locatingeditingfactualassociations} is a more challenging dataset designed to compare counterfactual and factual statements. Each record in the dataset is derived from corresponding entries in PARAREL \cite{elazar2021measuring}, where all subjects, relations, and objects are entities from WikiData. Out-of-scope data is constructed by replacing subject entities with approximate entities that share the same predicate. The Counterfact dataset uses metrics similar to those of ZsRE to evaluate efficacy, generalization, and specificity.
\end{itemize}

\subsection{Metrics}\label{sec:metrics}

In this section we introduce the evaluation metrics used in this paper. Each metric is defined given a LLM \(M\), a knowledge fact prompt \((s_i, r_i)\), rephrased prompts \(N((s_i, r_i))\), unrelated prompts \(O((s_i, r_i))\), an edited target output \(o_i\), and the model’s original output \(o_i^c\):
\begin{itemize}
    \item \textbf{Efficacy\textsubscript{top}}: The proportion of cases where \(o_i\) has the highest predicted probability with knowledge fact prompt \((s_i, r_i)\).
    \begin{equation}
        \mathbf{E}_i \left[ o_i = \arg\max_o \mathbf{P}_{M} \left[ o \mid \left( s_i, r_i \right) \right] \right]
    \end{equation}
    
    \item \textbf{Generalization\textsubscript{top}}: The proportion of cases where \(o_i\) has the highest predicted probability with rephrased prompts \(N((s_i, r_i))\).
    \begin{equation}
        \mathbf{E}_i \left[ o_i = \arg\max_o \mathbf{P}_{M} \left[ o \mid N \left( \left( s_i, r_i \right) \right) \right] \right]
    \end{equation}
    
    \item \textbf{Specificity\textsubscript{top}}: The proportion of cases where \(o_i^c\) has the highest predicted probability with unrelated prompts \(O((s_i, r_i))\).
    \begin{equation}
        \mathbf{E}_i \left[ o_i^c = \arg\max_o \mathbf{P}_{M} \left[ o \mid O \left( \left( s_i, r_i \right) \right) \right]  \right]
    \end{equation}

    \item \textbf{Efficacy\textsubscript{larger}}: The proportion of cases where \(o_i\) is more probable than \(o_i^c\) with knowledge fact prompt \((s_i, r_i)\).
    \begin{equation}
        \mathbf{E}_i \left[ \mathbf{P}_{M} \left[ o_i \mid  \left( s_i, r_i \right) \right] > \mathbf{P}_{M} \left[ o_c^i \mid  \left( s_i, r_i  \right) \right] \right].
    \end{equation}

    \item \textbf{Generalization\textsubscript{larger}}: The proportion of cases where \(o_i\) is more probable than \(o_i^c\) with rephrased prompts \(N((s_i, r_i))\).
    \begin{equation}
        \mathbf{E}_i \left[ \mathbf{P}_{M} \left[ o_i \mid N\left( \left( s_i, r_i \right) \right) \right] > \mathbf{P}_{M} \left[ o_c^i \mid N\left( \left( s_i, r_i \right) \right) \right] \right].
    \end{equation}

    \item \textbf{Specificity\textsubscript{larger}}: The proportion of cases where \(o_i^c\) is more probable than \(o_i\) with unrelated prompts \(O((s_i, r_i))\).
    \begin{equation}
        \mathbf{E}_i \left[ \mathbf{P}_{M} \left[ o_i^c \mid O\left( \left( s_i, r_i \right) \right) \right] > \mathbf{P}_{M} \left[ o_c \mid O\left( \left( s_i, r_i \right) \right) \right] \right].
    \end{equation}
    
\end{itemize}

When evaluating the metrics on the CounterFact dataset, to better verify the editing effectiveness, we did not use the \(o_i^c\) provided in the dataset. Instead, we recalculated the actual \(o_i^c\) from the model.

\subsection{Implementation Details} \label{sec:details}

We implement DeltaEdit on GPT2-XL and Llama3-8B, following the configuration described in AlphaEdit \cite{fang2024alphaedit}. All experiments are conducted on Linux system and implemented using PyTorch. The details are as follows:

\textbf{GPT2-XL}: The critical layers for editing are [13, 14, 15, 16, 17]. During the training of \(\alpha\), we perform 20 optimization steps, with a learning rate of 0.5, a sliding average coefficient \(\delta\) of 0.9, and a hyperparameter $\eta$ of 3. The experiments are conducted on a single RTX 4090 (24GB).

\textbf{Llama3-8B}: The critical layers for editing are [4, 5, 6, 7, 8]. During the training of \(\alpha\), we perform 25 optimization steps, with a learning rate of 0.1, a sliding average coefficient \(\delta\) of 0.9, and a hyperparameter $\eta$ of 1.5. The experiments are conducted on a single A100 (40GB).

The curves in Figure 2, Figure 3 and Figure 4 of the main paper are derived from the final edited layers of the models, specifically the 17th layer of GPT2-XL and the 8th layer of Llama3-8B, which are the layers most significantly impacted by the edits.

To enhance the experimental challenge, the batch size for each edit is set to 1 in all experiments in this paper. Our experimental results in Table 1 of the main paper are the average values obtained from three repeated experiments.

\subsection{Baselines}
In this study, we introduce five baseline models. For the hyperparameter settings of the baseline methods, except those mentioned in Appendix \ref{sec:details}, we reproduced the methods using the original code provided in their respective papers.

\textbf{ROME}: ROME is a method designed to update specific factual associations in large language models. By identifying key neuron activations in the feed-forward networks of middle layers that influence factual predictions, ROME directly modifies the weights of feedforward layers to edit these associations. ROME demonstrates that the feedforward modules in intermediate layers play a critical role in storing and recalling factual knowledge, making direct manipulation of the model a feasible editing technique.

\textbf{MEMIT}: MEMIT is a scalable multilayer update algorithm designed to efficiently insert new factual memories into transformer-based language models. Building on the direct editing approach of ROME, MEMIT targets specific weights in transformer modules that act as causal mediators for factual knowledge recall. This approach allows MEMIT to update thousands of new factual associations within the model.

\textbf{PRUNE}: PRUNE is a model editing framework designed to preserve the general capabilities of large language models during successive edits. PRUNE addresses the degradation of model performance caused by an increasing number of edits by applying a condition number constraint to the editing matrix, thereby limiting disruptions to the knowledge stored in the model. By controlling the numerical sensitivity of the model, PRUNE ensures that its overall ability remains intact while performing edits.

\textbf{RECT}: RECT is a method aimed at mitigating the unintended side effects of model editing on the general capabilities of large language models. While model editing can improve the factual accuracy of a model, it often reduces its performance on tasks such as reasoning and question answering. RECT prevents overfitting by regularizing weight updates during the editing process, avoiding excessive modifications. This enables RECT to maintain the model's general capabilities while achieving high editing performance.

\textbf{AlphaEdit}: AlphaEdit is a simple and efficient method designed to improve the editing performance of large language models. By introducing a null-space constraint, AlphaEdit projects the update results into the null space that preserves existing knowledge, preventing the issue of hidden representation distribution shifts caused by overfitting to updated knowledge. This approach effectively reduces update errors while maintaining the accuracy of existing knowledge, alleviating problems of model forgetting and collapse.

\newpage

\section{Algorithm}\label{sec:alg}

Algorithm \ref{alg:deltaedit} demonstrates the execution process of DeltaEdit. There is one implementation detail not mentioned in the main text: to prevent the mean $m$ from growing excessively, outliers are avoided during the update process. Specifically, the mean $m$ and variance $v$ are updated only when the value of $\|\Delta_{\text{history}} k_e\|_2^2$ falls within a reasonable range.

\begin{algorithm} 
\caption{DeltaEdit}  \label{alg:deltaedit}
\label{alg1}  
\begin{algorithmic}[1] 
\REQUIRE Sequence of edits $\mathbb{E}_T = \{\mathcal{E}_1, \mathcal{E}_2, \dots, \mathcal{E}_T\}$, original weight $W$, hyperparameter $\eta$, sliding average coefficient $\delta$, null space of unrelated input representations $\mathbb{P}$
\ENSURE Edited parameters $\widetilde{W}$  
\STATE Initialize $\Delta_{\text{history}} \leftarrow 0$, $m \leftarrow 0$, $v \leftarrow 0$, $K_p K_p^\top \leftarrow 0$, $\alpha_e \leftarrow \mathbf{0}$
\FOR{$\mathcal{E}_e \in \mathbb{E}_T$}
    \STATE Obtain the input representation $k_e$
    \STATE Determine whether to execute the orthogonal strategy, the first 5 edits skip to initializing \( m \) and \( v \):
    \IF{$\|\Delta_{\text{history}} k_e\|_2^2 > m + \eta \sqrt{v} \text{ \textbf{and} } e \geq 5 $}
        \STATE Construct the column space matrix:
        \STATE $D \leftarrow \Delta_{\text{history}} \Delta_{\text{history}}^\top$
        \STATE Perform Singular Value Decomposition (SVD) on $D$:
        \STATE $\{U, \Lambda, U^\top\} \leftarrow \text{SVD}(D)$
        \STATE Extract non-zero eigenvalues and corresponding eigenvectors:
        \STATE $N \leftarrow \{\text{non-zero eigenvalues of } \Lambda\}$
        \IF{Number of eigenvalues in $N$ exceeds $3/4$ of $\dim(\alpha_e)$}
            \STATE Remove smallest eigenvalues from $N$ until $|N| \leftarrow 3/4 \times \dim(\alpha_e)$
        \ENDIF
        \STATE Select eigenvectors corresponding to remaining eigenvalues in $N$ to form $\hat{U}$
        \STATE Compute the null space projection matrix:
        \STATE $P \leftarrow I - \hat{U}\hat{U}^\top$
        \ELSE
        \STATE $P \leftarrow I$
        \STATE Update the mean \(m\) and variance \(v\) only when the value of $\|\Delta_{\text{history}} k_e\|_2^2$ falls within a reasonable range:
        \STATE $m \leftarrow \delta m + (1 - \delta) \| \Delta_{\text{history}} k_e \|_2^2$
        \STATE $v \leftarrow \delta v + (1 - \delta) \left( \| \Delta_{\text{history}} k_e \|_2^2 - m \right)^2$
    \ENDIF
    \STATE Compute $\alpha_e:$
    \FOR{$\text{n} \in \text{num\_teps}$}
        \STATE Update $\alpha_e$ by optimizing loss function 
        \STATE $\alpha_e \leftarrow P \alpha_e$
    \ENDFOR
    \STATE Compute $\beta_e:$
    \STATE $\beta_e \leftarrow \left(\mathbb{P} K_p K_p^\top + \mathbb{P} k_e k_e^\top + I \right)^{-1} \mathbb{P} k_e$
    \STATE Update parameters:
    \STATE $W \leftarrow W + \alpha_e \beta_e^\top$, $\Delta_{\text{history}} \leftarrow \Delta_{\text{history}} + \alpha_e \beta_e^\top$, $K_p K_p^\top \leftarrow K_p K_p^\top + k_e k_e^\top$
\ENDFOR
\STATE \textbf{return} Updated weight $\widetilde{W} \leftarrow W$
\end{algorithmic}  
\end{algorithm}

\newpage

\begin{figure*}[t!]
    \centering
    \includegraphics[width=\linewidth]{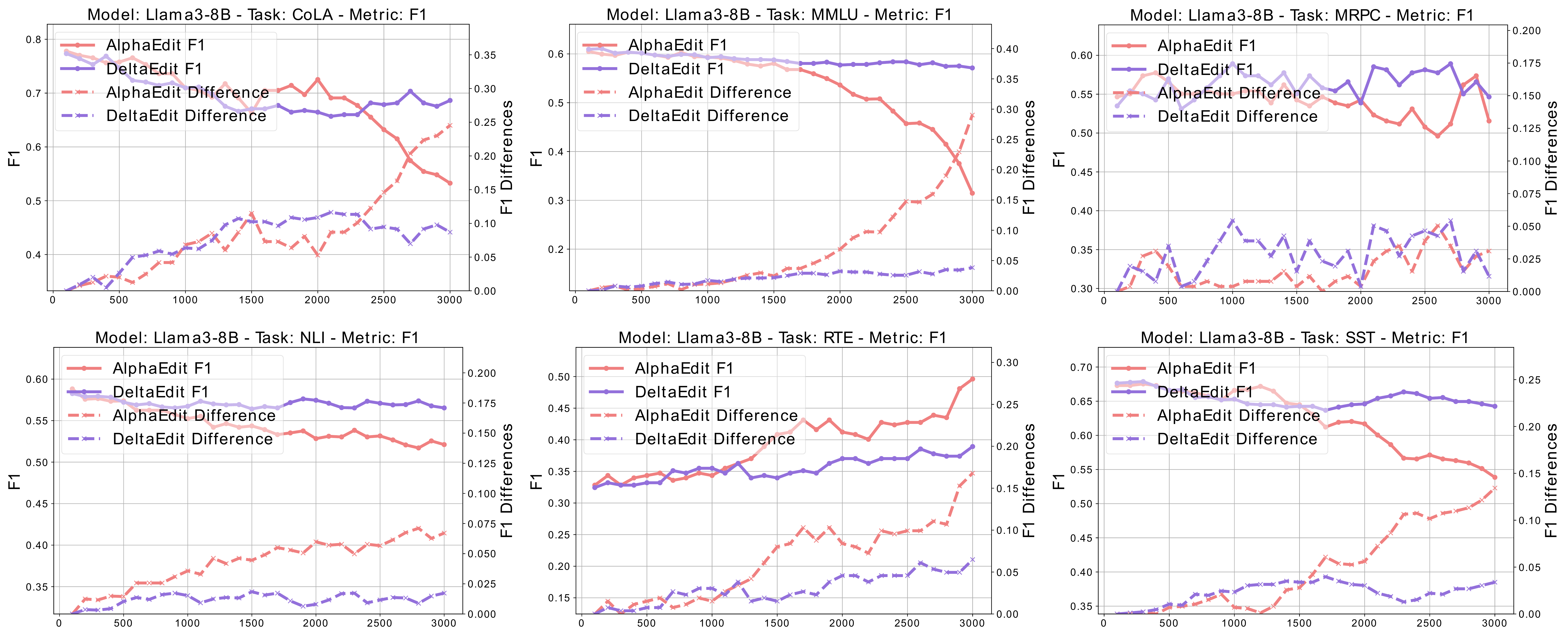}
    \caption{The solid lines represent the F1 scores of the edited LLaMA3-8B on six tasks (CoLA, MMLU, MRPC, NLI, RTE, and SST) as the number of edits increases on CounterFact. The dashed lines represent the differences between the F1 scores of the edited and pre-edit models over the edits.}
    \label{fig:general}
\end{figure*}

\begin{figure*}[htb!]
    \centering
    \includegraphics[width=\linewidth]{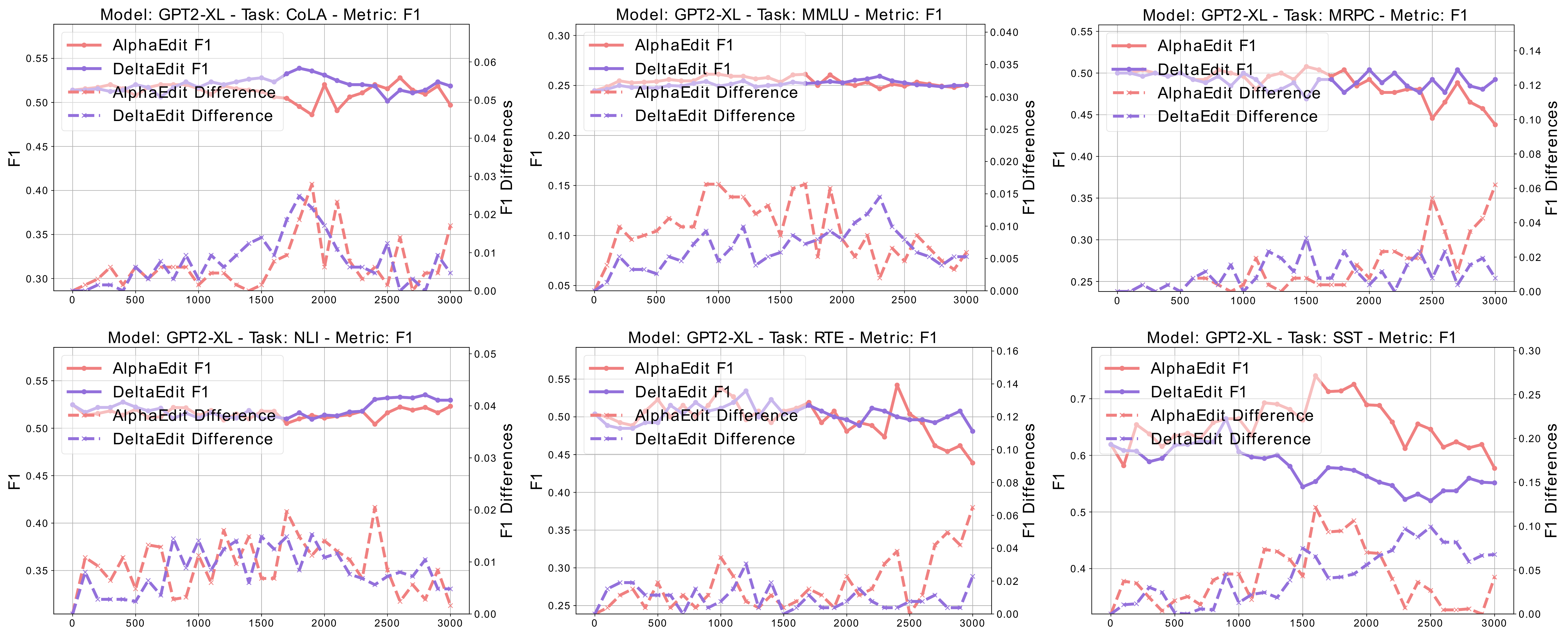}
    \caption{The solid lines represent the F1 scores of the edited GPT2-XL on six tasks (CoLA, MMLU, MRPC, NLI, RTE, and SST) as the number of edits increases. The dashed lines represent the differences between the F1 scores of the edited and pre-edit models over the edits.} \label{fig:gpt-general}
\end{figure*}

\section{Supplementary Experiments}

\subsection{General Capability Tests}\label{sec:general-test}

To evaluate the impact on the general capability of the edited models, we conduct tests on six tasks from the General Language Understanding Evaluation (GLUE) benchmark\cite{wang2018glue}: CoLA\cite{warstadt2019neural}, MMLU\cite{hendrycks2020measuring}, MRPC\cite{dolan2005automatically}, NLI\cite{williams2017broad}, RTE\cite{bentivogli2009fifth}, and SST\cite{socher2013recursive}. The experimental results on Llama3-8B are shown in Figure \ref{fig:general}, which depicts the F1 scores of the models and their corresponding F1 differences (i.e., the difference between the F1 scores of the edited and original models) as the number of edits increases on CounterFact.

The results demonstrate that DeltaEdit consistently preserves the original model performance in all metrics. Specifically, on tasks such as CoLA, MMLU, NLI, and RTE, DeltaEdit exhibits smaller F1 differences compared to AlphaEdit. In MRPC, while DeltaEdit exhibits some fluctuations in the F1 difference, the maximum deviation remains as small as 0.054, indicating that the edits have only a minor impact on performance. On the RTE task, a notable phenomenon occurs: AlphaEdit leads to an increase in F1 scores. However, this improvement does not indicate a positive outcome, as it reflects an unintended interference with the general capability of the model. These results collectively highlight that DeltaEdit effectively integrates the required edits while successfully maintaining the model's original capabilities.

Figure \ref{fig:gpt-general} demonstrates the test results of GPT2-XL on six tasks, edited using AlphaEdit and DeltaEdit on CounterFact. It is evident that GPT2-XL has minimal ability to perform these tasks. Without any edits, GPT2-XL's performance on five tasks (excluding SST) is close to random guessing, resulting in minimal performance changes after editing. On SST, although the performance after 3,000 edits with DeltaEdit is lower than that with AlphaEdit, AlphaEdit appears to cause greater performance fluctuations compared to DeltaEdit.

\subsection{Supplementary experimental results of Llama3-8B} \label{sec:sup-llama}
In this section, we present the experimental results for editing Llama3-8B that are omitted in Section~3.3 and Section~3.4 due to excessively large numerical values. 

Table \ref{tab:llama3-noise} reports the \(noise_E\) values after applying MEMIT to edit Llama3-8B across different numbers of edits on CounterFact dataset. As shown, \(noise_E\) increases rapidly to extremely large values, which explains MEMIT's poor performance in Table 1 of the main paper. It is worth noting that while MEMIT achieves the highest Specificity\textsubscript{larger} after editing Llama3-8B, this is primarily because most of its edits fail, resulting in little to no change in the output probability of the target tokens.

Table \ref{tab:k_beta} presents the values of \( k^\top \beta \) after applying MEMIT and AlphaEdit to edit Llama3-8B on CounterFact dataset across different numbers of edits. When using AlphaEdit, the trend of \( k^\top \beta \) on Llama3-8B follows a similar downward trajectory as observed on GPT-2. This occurs because the differences between the \(\beta\) vectors computed by AlphaEdit are substantial, and as the number of edits increases, the denominator in the averaging process (i.e., the number of edits) grows, leading to this decline. Moreover, the \( k^\top \beta \) values achieved by MEMIT are significantly higher than those produced by AlphaEdit.

\begin{table}[htb!]
    \centering
    \caption{The \(noise_E\) after applying MEMIT to edit Llama3-8B on CounterFact across different numbers of edits.} \label{tab:llama3-noise}
    \begin{tabular}{p{2.5cm}*{6}{p{1.13cm}}p{1.13cm}}
    \hline
        Number of edits & 200 & 500 & 1000 & 1500 & 2000 & 2500 & 3000 \\ \hline
        MEMIT-\(noise_E\) & 1.79 & 2780.7 & 20369.8 & 51434.5 & 73090.2 & 92678.4 & 111939.2 \\ \hline
    \end{tabular}
\end{table}

\begin{table}[htb!]
    \centering
    \caption{The values of \( k^\top \beta \) after applying MEMIT and AlphaEdit to edit Llama3-8B on CounterFact across different numbers of edits.}\label{tab:k_beta}
    \begin{tabular}{p{2.5cm}*{6}{p{1.13cm}}p{1.13cm}}
    \hline
        Number of edits & 200 & 500 & 1000 & 1500 & 2000 & 2500 & 3000 \\ \hline
        MEMIT-\( k^\top \beta \) & 0.0215 & 0.0323 & 0.2001 & 0.3210 & 0.3961 & 0.4458 & 0.4809 \\ \hline
        AlphaEdit-\( k^\top \beta \) & 0.0100 & 0.0064 & 0.0044 & 0.0035 & 0.0030 & 0.0026 & 0.0023 \\ \hline
    \end{tabular}
\end{table}

\begin{figure*}[tb!]
    \centering
    \begin{subfigure}[t]{0.31\textwidth}
        \centering
        \includegraphics[width=\linewidth]{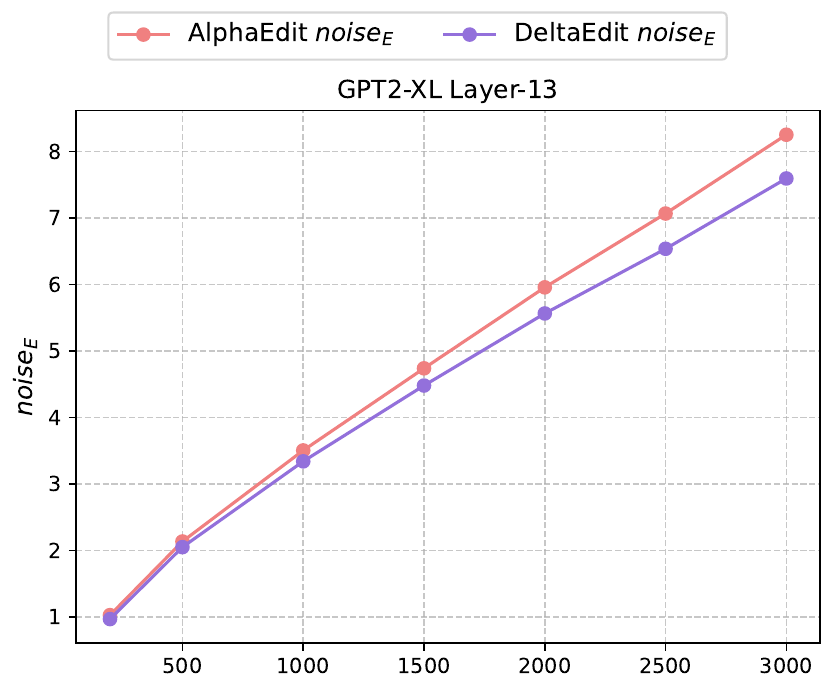}
        \caption{} 
    \end{subfigure}
    \begin{subfigure}[t]{0.31\textwidth}
        \centering
        \includegraphics[width=\linewidth]{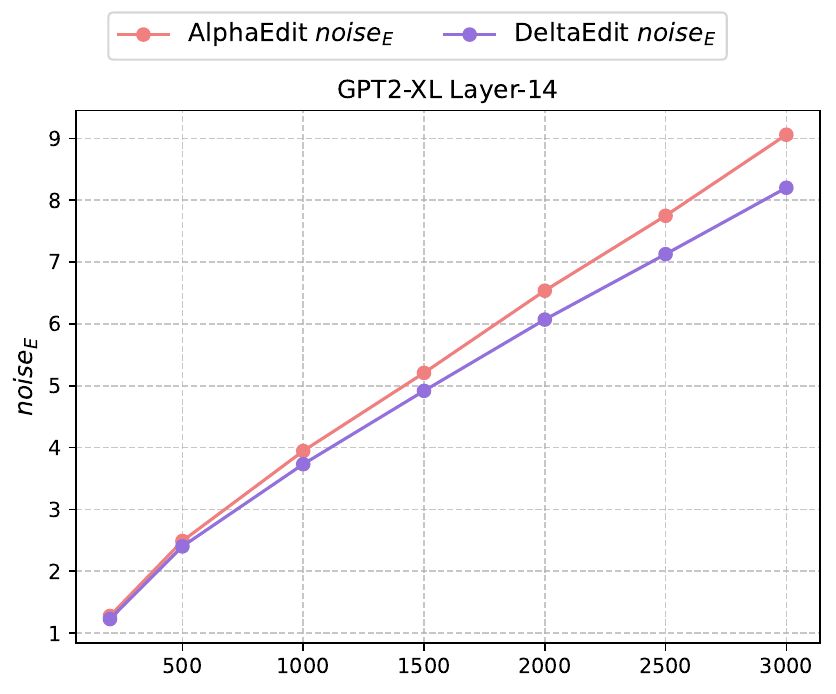}
        \caption{}
    \end{subfigure}
    \begin{subfigure}[t]{0.31\textwidth}
        \centering
        \includegraphics[width=\linewidth]{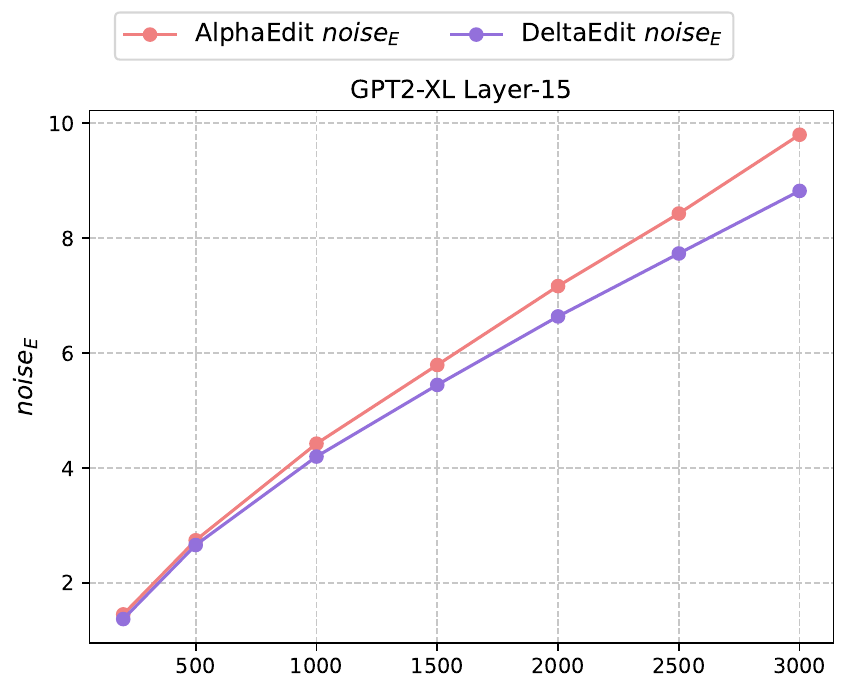}
        \caption{}
    \end{subfigure}

    \vspace{-10pt}

    \begin{subfigure}[t]{0.31\textwidth}
        \centering
        \includegraphics[width=\linewidth]{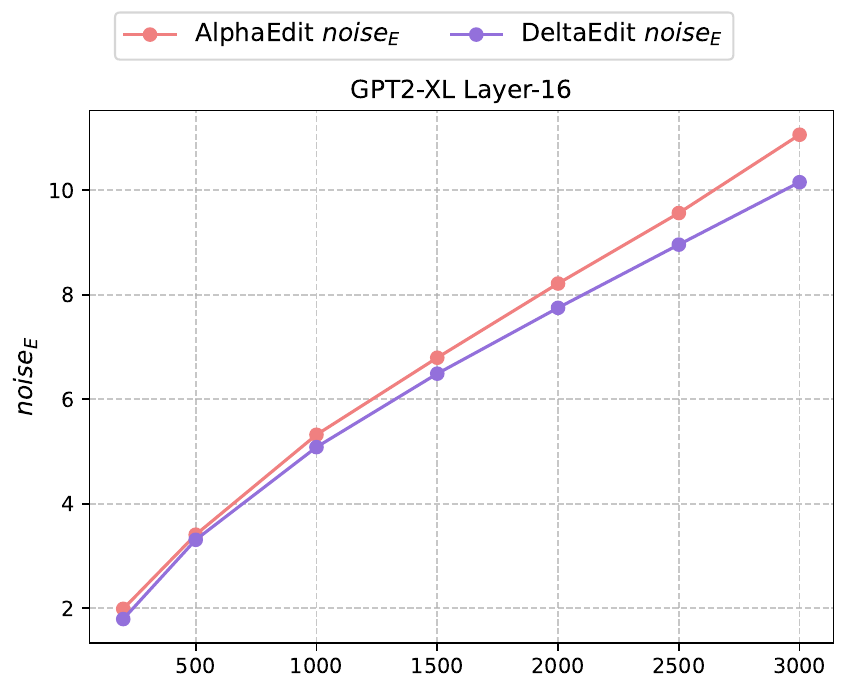}
        \caption{}
    \end{subfigure}
    \begin{subfigure}[t]{0.31\textwidth}
        \centering
        \includegraphics[width=\linewidth]{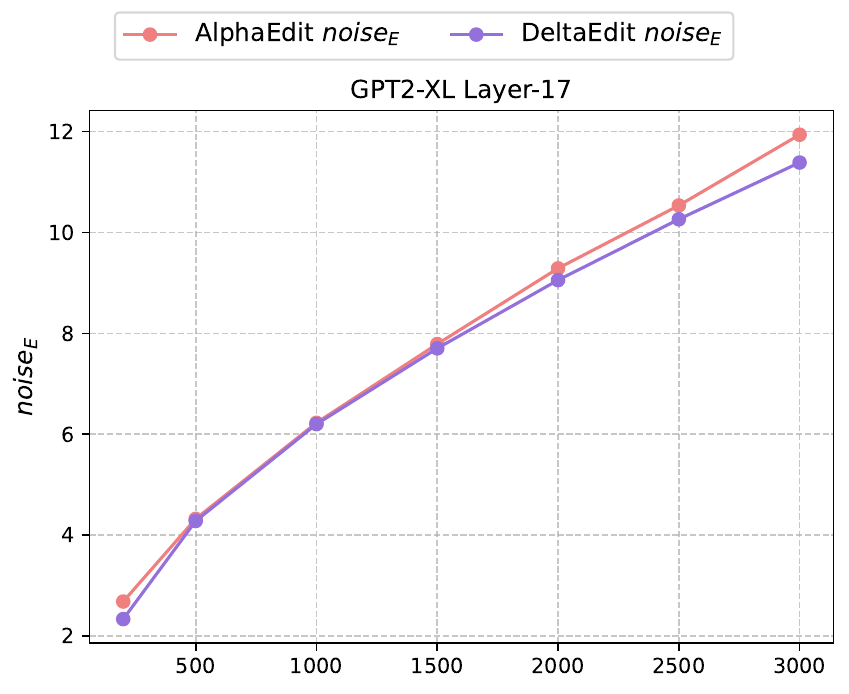}
        \caption{}
    \end{subfigure}
    \vspace{-20pt}
    \caption{The \(noise_E\) after applying AlphaEdit and DeltaEdit to edit GPT2-XL on CounterFact across different numbers of edits.} \label{fig:gpt2-noise-all}
\end{figure*}

\begin{figure*}[tb!]
    \centering
    \begin{subfigure}[t]{0.31\textwidth}
        \centering
        \includegraphics[width=\linewidth]{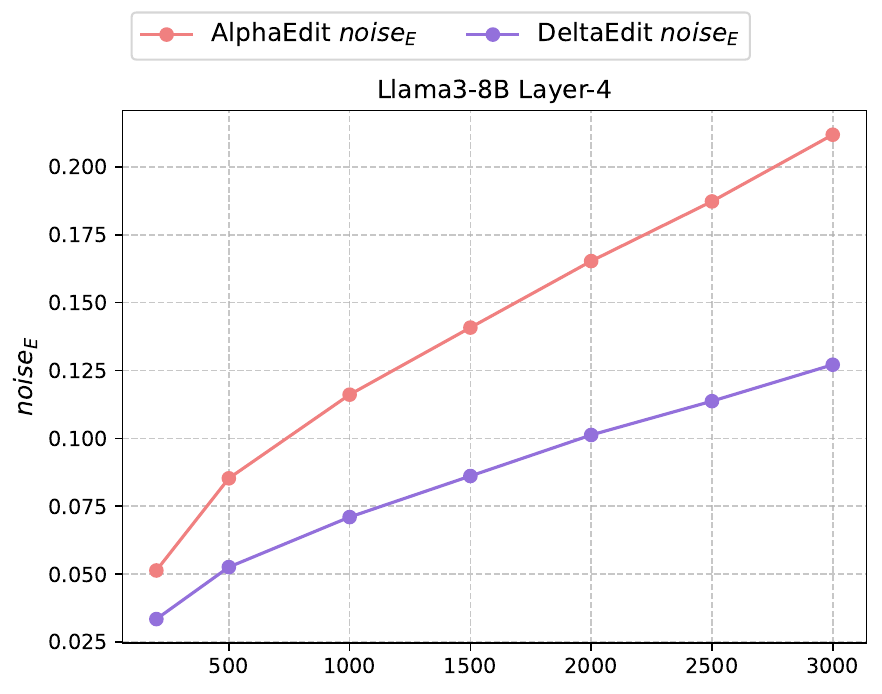}
        \caption{}
    \end{subfigure}
    \begin{subfigure}[t]{0.31\textwidth}
        \centering
        \includegraphics[width=\linewidth]{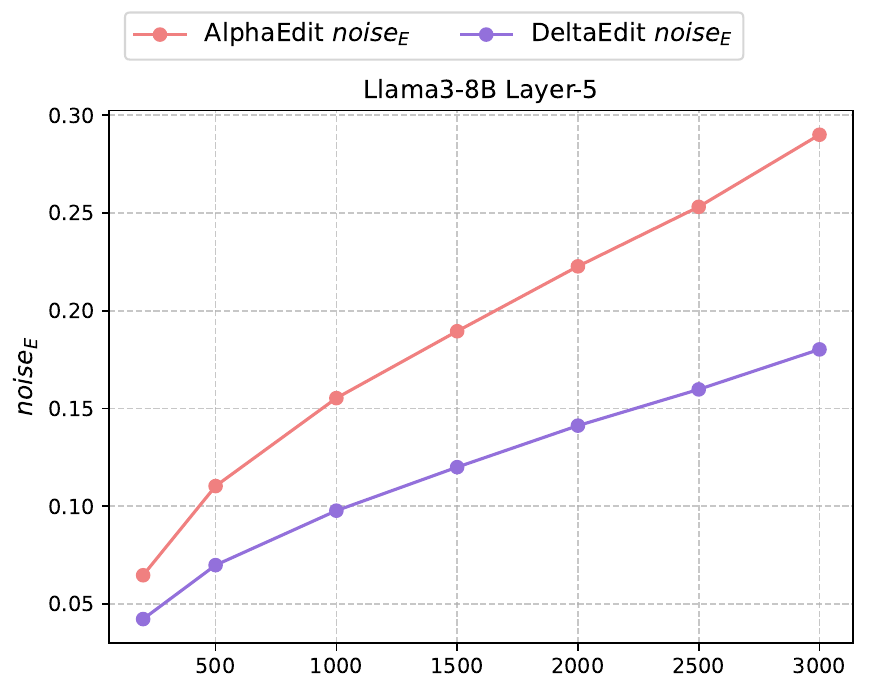}
        \caption{}
    \end{subfigure}
    \begin{subfigure}[t]{0.31\textwidth}
        \centering
        \includegraphics[width=\linewidth]{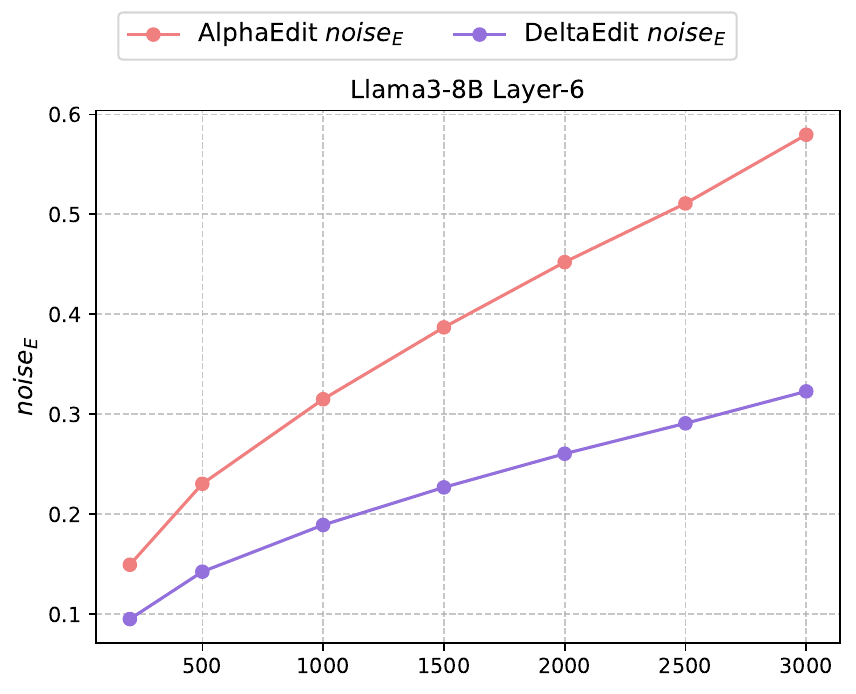}
        \caption{}
    \end{subfigure}

    \vspace{-10pt}

    \begin{subfigure}[t]{0.31\textwidth}
        \centering
        \includegraphics[width=\linewidth]{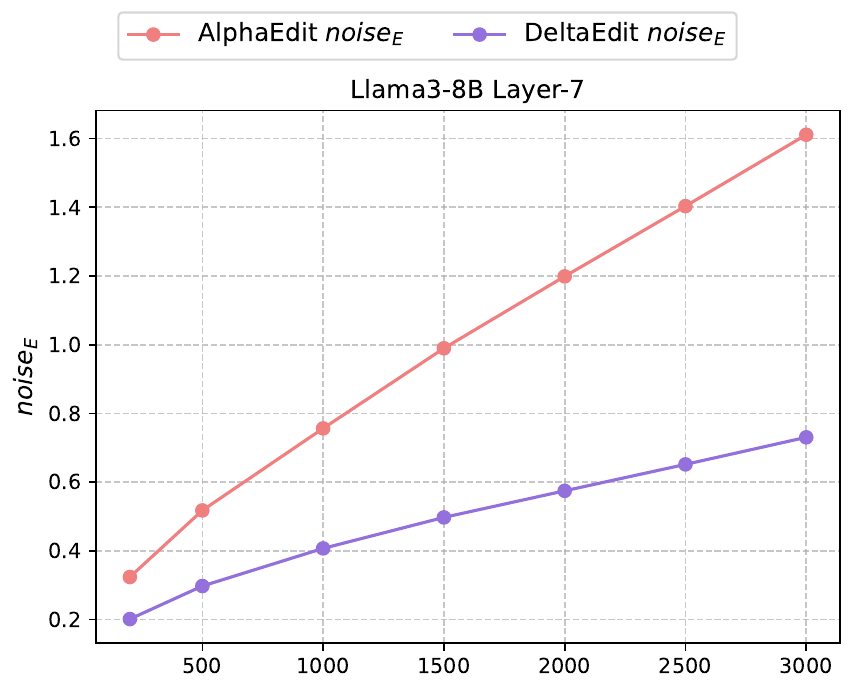}
        \caption{}
    \end{subfigure}
    \begin{subfigure}[t]{0.31\textwidth}
        \centering
        \includegraphics[width=\linewidth]{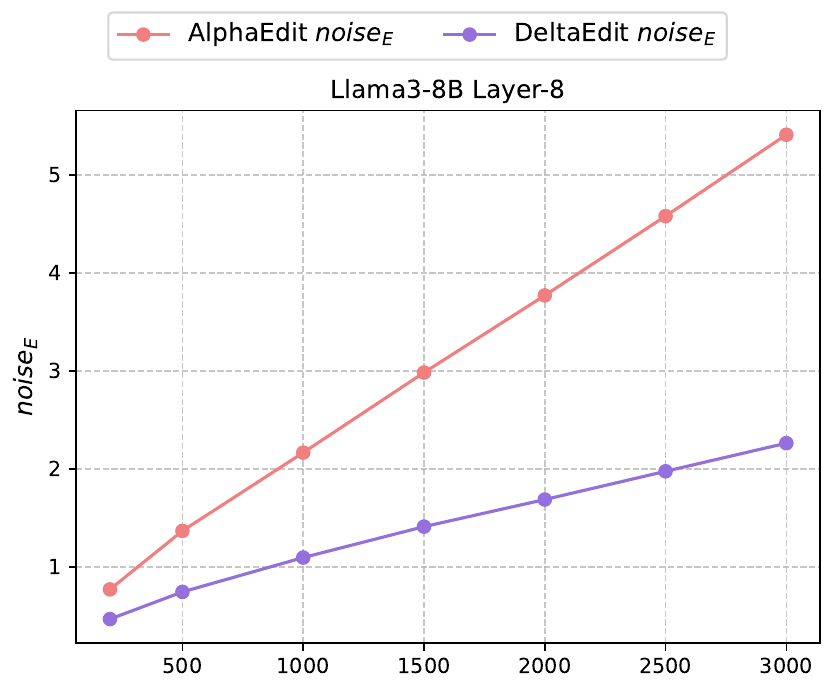}
        \caption{}
    \end{subfigure}
    \vspace{-20pt}
    \caption{The \(noise_E\) after applying AlphaEdit and DeltaEdit to edit Llama3-8B on CounterFact across different numbers of edits.}\label{fig:llama3-noise-all}
\end{figure*}

\subsection{The Impact of DeltaEdit on noise\textsubscript{E} Across All Edited Layers} \label{sec:all-layer}
Figures \ref{fig:gpt2-noise-all} and \ref{fig:llama3-noise-all} show the \(noise_E\) in the multi-layer editing setup\cite{meng2023masseditingmemorytransformer,fang2024alphaedit} after applying AlphaEdit and DeltaEdit to edit GPT2-XL and Llama3-8B on CounterFact across different numbers of edits. Experimental results show that DeltaEdit can effectively reduce the \(noise_E\) across all edited layers.

\subsection{Analysis experiments of GPT2-XL}
In this section, we present the results of the hidden representations analysis of GPT2-XL. 

\begin{figure}[htb!]
    \centering
    \includegraphics[width=0.5\textwidth]{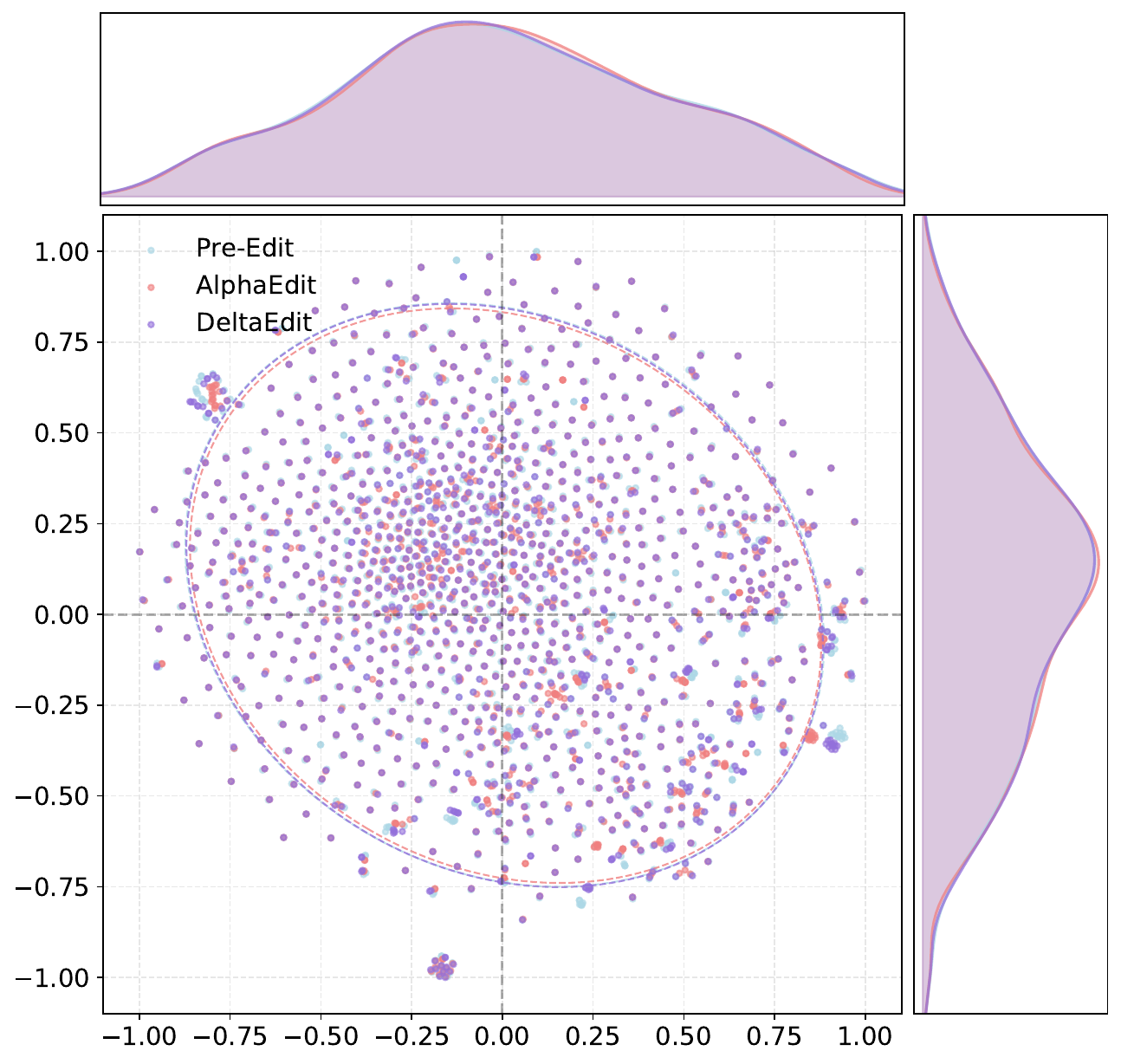} 
    \caption{The distribution of hidden representations of pre-edited and post-edited GPT2-XL after dimensionality reduction. The top and right curve graphs display the marginal distributions for two reduced dimensions. The dashed lines represent the 0.95 confidence intervals.}
    \label{fig:gpt2-distribution}
\end{figure}

Figure \ref{fig:gpt2-distribution} presents the t-SNE dimensionality reduction results of GPT2-XL's hidden representations after 3,000 edits using AlphaEdit and DeltaEdit on CounterFact. It is evident that neither AlphaEdit nor DeltaEdit significantly alters the distribution of the hidden representations. 

\subsection{Impact of the Hyperparameter $\eta$}

\begin{figure*}[htb!]
    \centering
    \begin{subfigure}[t]{0.45\textwidth} 
        \centering
        \includegraphics[width=\linewidth]{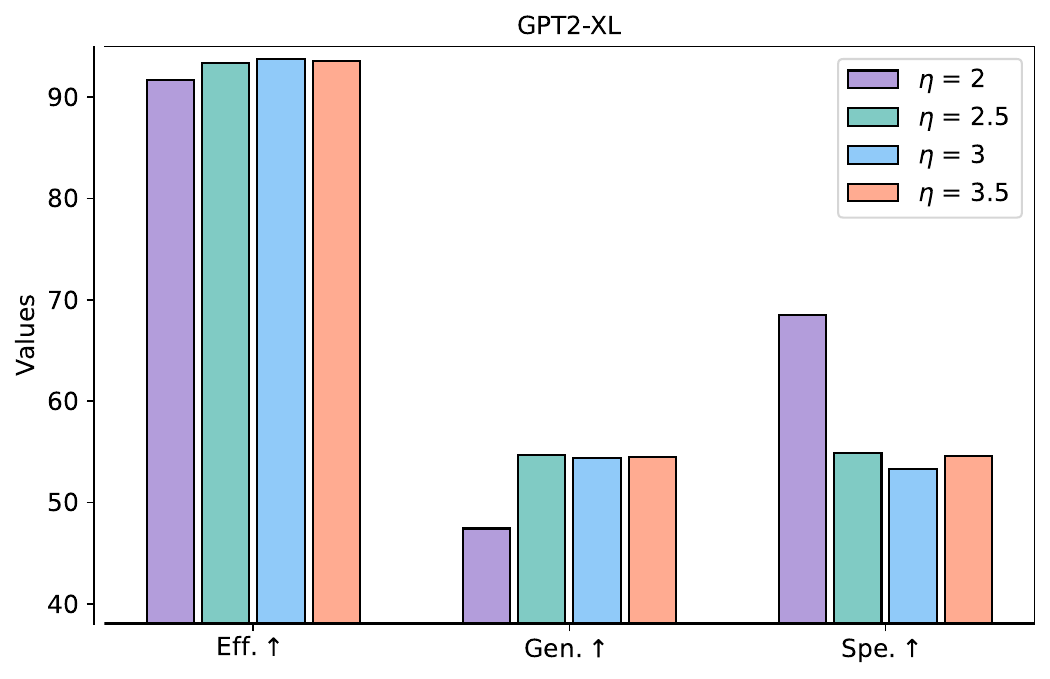}
        \caption{}
    \end{subfigure}\hfill
    \begin{subfigure}[t]{0.45\textwidth}
        \centering
        \includegraphics[width=\linewidth]{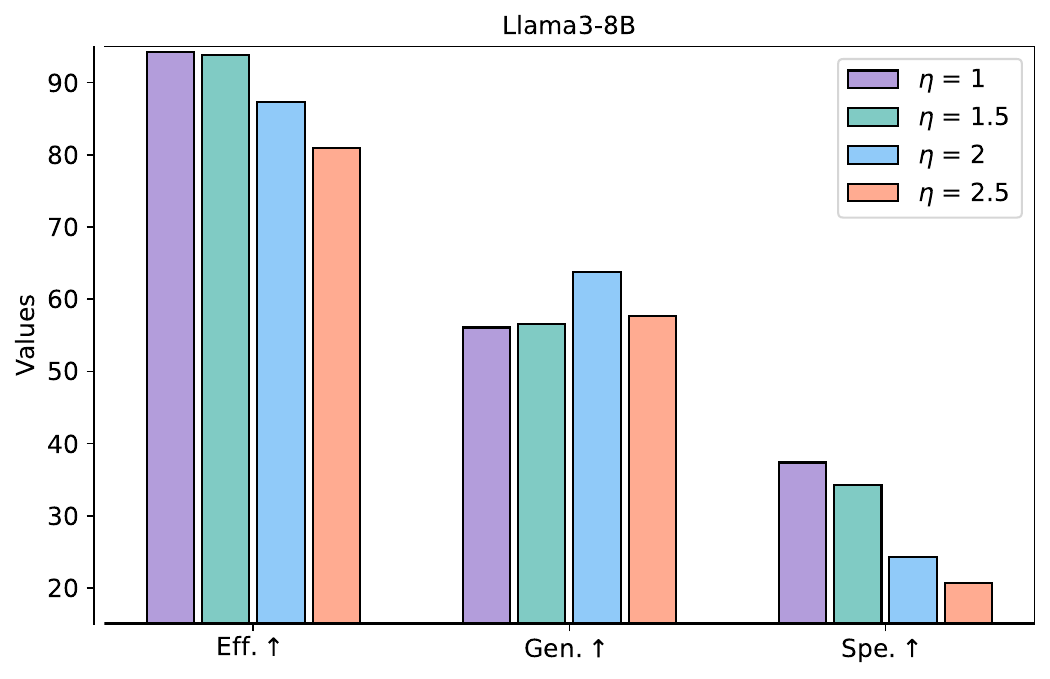}
        \caption{}
    \end{subfigure}
    \caption{Eff., Gen., and Spe. denote Efficacy\textsubscript{top}, Generalization\textsubscript{top}, and Specificity\textsubscript{top}. The figure shows the test results after 3,000 edits on CounterFact using different hyperparameters $\eta$.} \label{fig:eta}
\end{figure*}

In this section, we demonstrate the impact of the hyperparameter $\eta$. Figure~\ref{fig:eta} presents the test results after 3,000 edits on CounterFact using varying hyperparameter values of $\eta$. On GPT2-XL, $\text{Efficacy}_{\text{top}}$ shows minimal variation across different $\eta$ values. However, $\text{Generalization}_{\text{top}}$ rises significantly as $\eta$ increases from 2 to 2.5, while $\text{Specificity}_{\text{top}}$ drops sharply. Beyond $\eta = 2.5$, these metrics stabilize. On Llama3-8B, as $\eta$ increases, both $\text{Efficacy}_{\text{top}}$ and $\text{Specificity}_{\text{top}}$ decrease, while $\text{Generalization}_{\text{top}}$ is notably higher at $\eta = 2$ compared to other values. These results suggest that increasing $\eta$ beyond a certain threshold significantly affects editing performance: $\text{Generalization}_{\text{top}}$ improves, while $\text{Specificity}_{\text{top}}$ declines. Interestingly, $\text{Efficacy}_{\text{top}}$ exhibits contrasting trends across models: it increases on GPT2-XL but decreases on Llama3-8B. We hypothesize that this difference arises because small changes in $\eta$ lead to significant variations in the number of orthogonal strategy executions, which impacts performance. Additionally, the hidden representations of GPT2-XL have a lower dimensionality compared to those of Llama3-8B. This suggests that for GPT2-XL, excessive orthogonal intensity caused by small $\eta$ values may degrade editing performance. Above all, it is necessary to find an appropriate $\eta$ to achieve a balance between editing effectiveness, generalization capability, and the retention of other knowledge.

\subsection{Case Study}\label{sec:casestudy}
We select several cases from CounterFact to evaluate the generation capabilities of models after sequence editing. The outputs of GPT-2 XL are presented in Table \ref{tab:gpt-case}, while those of Llama3-8B are shown in Figure \ref{tab:llama-case}. It is observed that the model edited with MEMIT loses the ability to generate coherent sentences and instead resorts to word repetition. Although the model edited with AlphaEdit produces relatively fluent sentences, the generated content is unrelated to the edited knowledge, indicating a divergence between its outputs and the expected results after sequence editing. In contrast, DeltaEdit not only preserves the ability to generate fluent sentences but also ensures that the generated content aligns with the edited knowledge. As shown in these tables, when Danielle Darrieux's native language is modified, the edited model infers that she is born in a country where English is the primary language.

\vskip -1.5em
\begin{table}[!htp]
    \centering
    \caption{The outputs of GPT2-XL.} \label{tab:gpt-case}
    \begin{tabular}{p{2cm}|m{8cm}<{}}
        \hline
        \multicolumn{2}{c}{GPT2-XL} \\
        \hline
        \multicolumn{2}{c}{Edited Knowledge: The mother tongue of Danielle Darrieux is English} \\
        \hline
        \multicolumn{2}{c}{Prompt: Danielle Darrieux was born in} \\
        \hline
        Method & Output \\
        \hline
        Original & Danielle Darrieux was born in Montreal, Quebec, Canada, to a French-Canadian father and a French-Canadian mother. She is of French-Canadian, English, and Scottish descent.\\
        \hline
        MEMIT & Danielle Darrieux was born in Chicago, California Dutch, French French Dutch -- French Pope Dutch Dutch Dutch Dutch Dutch Dutch Dutch Dutch German Dutch Dutch Dutch Dutch Dutch Dutch Dutch\\
        \hline
        AlphaEdit & Danielle Darrieux was born in Germany, and has a German passport. Her father was born in Germany, and is a German citizen, and has a German passport. The family is a mix of German parents, and her father is German.\\
        \hline
        DeltaEdit & Danielle Darrieux was born in Detroit, Michigan, on January 26, 1957. She is a native of Detroit, Michigan, and graduated from the University of Michigan in Ann Arbor, Michigan.\\
        \hline
    \end{tabular}
    \vskip -1.5em
 \end{table}

\begin{table}[!htp]
    \centering
    \caption{The outputs of Llama3-8B} \label{tab:llama-case}
    \begin{tabular}{p{2cm}|m{8cm}<{}}
        \hline
        \multicolumn{2}{c}{Llama3-8B} \\
        \hline
        \multicolumn{2}{c}{Edited Knowledge: The mother tongue of Danielle Darrieux is English} \\
        \hline
        \multicolumn{2}{c}{Prompt: Danielle Darrieux was born in} \\
        \hline
        Method & Output \\
        \hline
        Original & Danielle Darrieux was born in 1917, and is one of the most famous French actresses of all time. Her film career began in the late 1930's. Her first film was Le Grand Jeu. In 1951 she starred alongside Yves Montand in the film La Belle Ame\'ricaine. \\
        \hline
        MEMIT & Danielle Darrieux was born in addCriterion addCriterion addCriterion addCriterion addCriterion addCriterion addCriterion,LOCATION addCriterionernel addCriterion addCriterionernel\\
        \hline
        AlphaEdit & Danielle Darrieux was born in 1936 in Moscow, Russia. She was born in Moscow, Russia. She was born in Russia. She was born in Russia, Moscow, on December 20, She was born in Moscow, Russia.\\
        \hline
        DeltaEdit & Danielle Darrieux was born in London, England. She attended school in France, New York and Switzerland, and then attended the University of the Pacific (based in Costa Rica) for one year.\\
        \hline
    \end{tabular}
    \vskip -1.5em
 \end{table}

\end{document}